\RequirePackage{fix-cm}
\documentclass[twocolumn]{svjour3}          
\smartqed  
\usepackage{graphicx}
\usepackage{natbib}
\usepackage{amsmath}
\usepackage{amssymb}
\usepackage{bbm}
\usepackage{bm}
\usepackage{booktabs}
\usepackage{array} 
\usepackage{blindtext}
\usepackage{comment}

\usepackage{pifont}
\newcommand{\cmark}{\ding{51}}

\usepackage[dvipsnames]{xcolor}
\usepackage{color, colortbl}
\definecolor{citecolor}{HTML}{0071bc}
\definecolor{tabhighlight}{HTML}{e5e5e5}
\usepackage[colorlinks,citecolor=citecolor]{hyperref}

\makeatletter
\renewcommand\paragraph{
  \@startsection{paragraph} 
  {4} 
  {\z@} 
  {.5em \@plus1ex \@minus.2ex} 
  {-.5em} 
  {\normalfont\normalsize\bfseries} 
}
\makeatother
\usepackage{algorithm}
\usepackage{amsfonts}
\usepackage[caption=false,font=normalsize,labelfont=sf,textfont=sf]{subfig}
\usepackage{textcomp}
\usepackage{stfloats}
\usepackage{url}
\usepackage{verbatim}
\usepackage{cite}
\usepackage{epsfig}
\usepackage{ifthen}
\usepackage{caption}
\usepackage{multicol}
\usepackage{multirow}
\usepackage{makecell}
\usepackage{arydshln}
\usepackage{mathtools}
\usepackage{algpseudocode}
\usepackage{xfrac}
\usepackage{appendix}

\newcommand{\ie}{\textit{i.e.}}
\newcommand{\eg}{\textit{e.g.}}
\newcommand{\etal}{\textit{et~al.}}
\newcommand{\rf}[1]{{\textbf{\color{red}{#1}}}} 
\newcommand{\bd}[1]{{\color{blue}{\underline{#1}}}} 

\begin{document}
\sloppy

\title{Exploiting Diffusion Prior for Real-World Image Super-Resolution 
}


\author{Jianyi Wang   \and
        Zongsheng Yue \and
        Shangchen Zhou \and
        Kelvin C.K. Chan \and
        Chen Change Loy
}


\institute{Jianyi Wang \at
              S-Lab, Nanyang Technological University, Singapore \\
              \email{jianyi001@ntu.edu.sg}
           \and
           Zongsheng Yue \at
              S-Lab, Nanyang Technological University, Singapore \\
              \email{zongsheng.yue@ntu.edu.sg}
           \and
           Shangchen Zhou \at
              S-Lab, Nanyang Technological University, Singapore \\
              \email{s200094@ntu.edu.sg}
           \and
           Kelvin C.K. Chan \at
              S-Lab, Nanyang Technological University, Singapore \\
              \email{chan0899@ntu.edu.sg}
           \and
           Chen Change Loy (Corresponding author) \at
              S-Lab, Nanyang Technological University, Singapore \\
              \email{ccloy@ntu.edu.sg}
}

\date{Received: date / Accepted: date}

\maketitle

\begin{abstract}
We present a novel approach to leverage prior knowledge encapsulated in pre-trained text-to-image diffusion models for blind super-resolution (SR). Specifically, by employing our time-aware encoder, we can achieve promising restoration results without altering the pre-trained synthesis model, thereby preserving the generative prior and minimizing training cost.
To remedy the loss of fidelity caused by the inherent stochasticity of diffusion models, we employ a controllable feature wrapping module that allows users to balance quality and fidelity by simply adjusting a scalar value during the inference process.
Moreover, we develop a progressive aggregation sampling strategy to overcome the fixed-size constraints of pre-trained diffusion models, enabling adaptation to resolutions of any size.
A comprehensive evaluation of our method using both synthetic and real-world benchmarks demonstrates its superiority over current state-of-the-art approaches.
Code and models are available at https://github.com/IceClear/StableSR.
\end{abstract}

\begin{figure*}[!t]
  \begin{center}
    \includegraphics[width=1.0\textwidth]{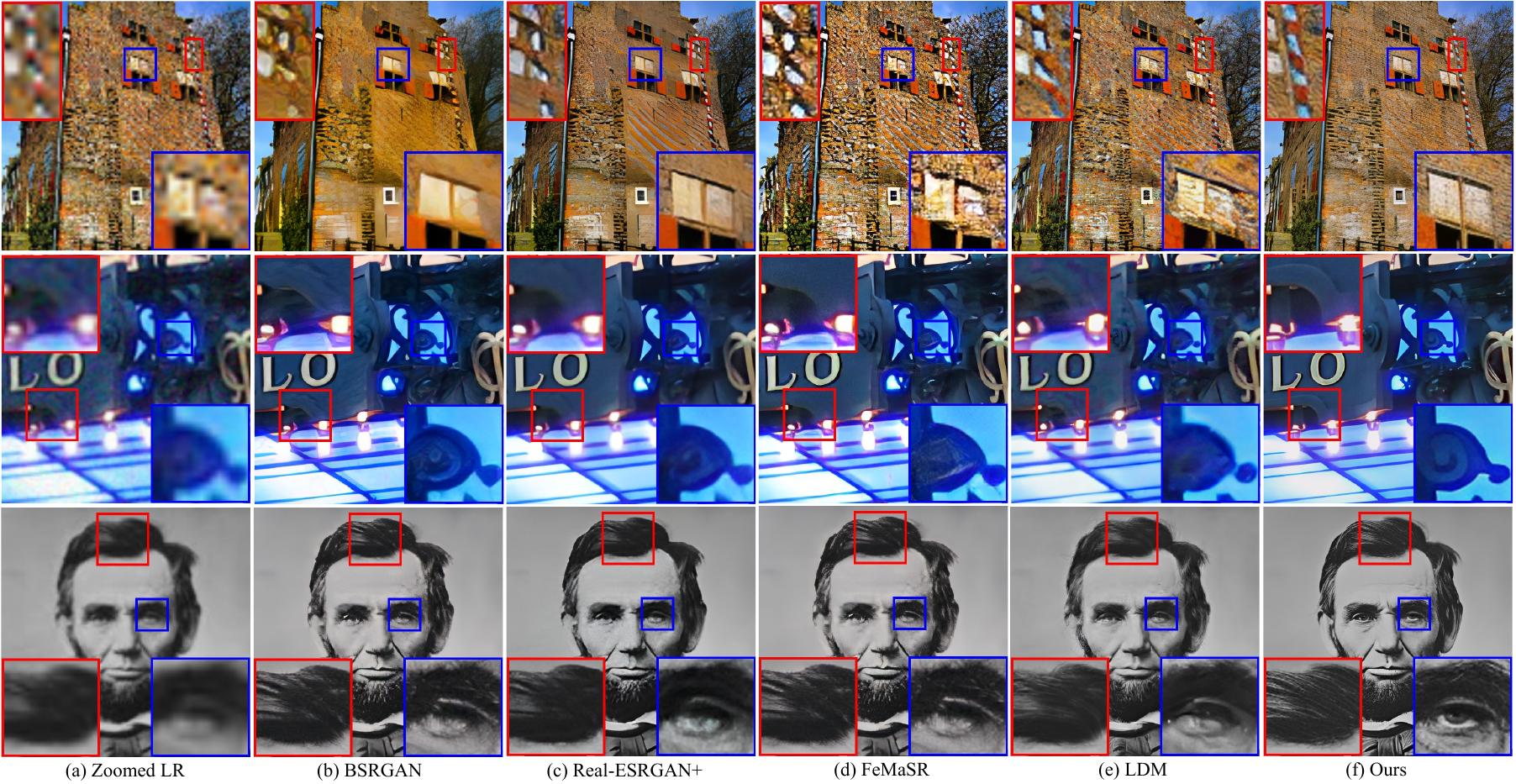}
    \caption{Qualitative comparisons of BSRGAN~\citep{zhang2021designing}, Real-ESRGAN+~\citep{wang2021realesrgan}, FeMaSR~\citep{chen2022femasr}, LDM~\citep{rombach2021highresolution}, and our StableSR on real-world examples. (\textbf{Zoom in for details})
    }
    \label{fig:fig1}
  \end{center}
  \vspace{-0.5cm}
\end{figure*}

\section{Introduction}

We have seen significant advancements in diffusion models~\citep{sohl2015deep,ho2020denoising,2021Score,nichol2022glide} for the task of image synthesis. Existing studies demonstrate that the diffusion prior, embedded in synthesis models like Stable Diffusion~\citep{rombach2021highresolution}, can be applied to various downstream content creation tasks, including image~\citep{choi2021ilvr,avrahami2022blended,hertz2022prompt,gu2022vector,mou2023t2i,zhang2023adding,gal2023designing} and video~\citep{wu2022tuneavideo,molad2023dreamix,qi2023fatezero} editing.
In this study, we extend the exploration beyond the realm of content creation and examine the potential benefits of using diffusion prior for super-resolution (SR). This low-level vision task presents an additional non-trivial challenge, as it requires high image fidelity in its generated content, which stands in contrast to the stochastic nature of diffusion models.

A common solution to the challenge above involves training a SR model from scratch~\citep{saharia2022image,rombach2021highresolution,sahak2023denoising,li2022srdiff}. To preserve fidelity, these methods use the low-resolution (LR) image as an additional input to constrain the output space. While these methods have achieved notable success, they often demand significant computational resources to train the diffusion model.
Moreover, training a network from scratch can potentially jeopardize the generative priors captured in synthesis models, leading to suboptimal performance in the final network.
These limitations have inspired an alternative approach~\citep{choi2021ilvr,wang2022zero,chungimproving,songpseudoinverse,meng2022diffusion}, which involves incorporating constraints into the reverse diffusion process of a pre-trained synthesis model.
This paradigm avoids the need for model training while leveraging the diffusion prior. However, designing these constraints assumes knowing the image degradations as a priori, which are typically unknown and complex. Consequently, such methods exhibit limited generalizability.

In this study, we present \textbf{StableSR}, an approach that \textbf{\textit{preserves pre-trained diffusion priors without making explicit assumptions about the degradations.}}
Specifically, unlike previous works~\citep{saharia2022image,rombach2021highresolution,sahak2023denoising,li2022srdiff} that concatenate the LR image to intermediate outputs, which requires one to train a diffusion model from scratch, our method only needs to fine-tune a lightweight \textit{time-aware encoder} and a few feature modulation layers for the SR task. 
When applying diffusion models for SR, the LR condition should provide adaptive guidance for each diffusion step during the restoration process, \ie, stronger guidance at earlier iterations to maintain fidelity and weaker guidance later to avoid introducing degradations.
To this end, our encoder incorporates a time embedding layer to generate time-aware features, allowing the features in the diffusion model to be adaptively modulated at different iterations.
Besides gaining improved training efficiency, keeping the original diffusion model frozen helps preserve the generative prior, which grants StableSR the capability of generating visually pleasant SR details and avoids overfitting to high-frequency degradations. 
Our experiments show that both the time-aware property of our encoder and the diffusion prior are crucial for achieving SR performance improvements.

To suppress randomness inherited from the diffusion model as well as the information loss due to the encoding process of the autoencoder~\citep{rombach2021highresolution}, inspired by Codeformer \citep{zhou2022codeformer}, we apply a \textit{controllable feature wrapping module} (CFW) with an adjustable coefficient to refine the outputs of the diffusion model during the decoding process of the autoencoder.
Unlike CodeFormer, the multiple-step sampling nature of diffusion models makes it hard to finetune the CFW module directly. 
We overcome this issue by first generating synthetic LR-HR pairs with the diffusion training stage. 
Then, we obtain the corresponding latent codes using our finetuned diffusion model given the LR images as conditions. 
In this way, CFW can be trained using the generated data.

Applying diffusion models to arbitrary resolutions has remained a persistent challenge, especially for the SR task. A simple solution would be to split the image into patches and process each independently. However, this method often leads to boundary discontinuity in the output.
To address this issue, we introduce a \textit{progressive aggregation sampling strategy}. Inspired by Jim\'{e}nez~\citep{jimenez2023mixture}, our approach involves dividing the image into overlapping patches and fusing these patches using a Gaussian kernel at each diffusion iteration. This process smooths out the boundaries, resulting in a more coherent output.
To avoid altering the output resolution of SR images, the overlapping sizes at the right and bottom boundaries are dynamically adjusted to fit the target resolution.

Adapting generative priors for real-world image super-resolution presents an intriguing yet challenging problem, and in this work, we offer a novel approach as a solution. We introduce a fine-tuning method that leverages pre-trained diffusion models without making explicit assumptions about degradations.
We address key challenges, such as fidelity and arbitrary resolution, by proposing simple yet effective modules. With our time-aware encoder, controllable feature wrapping module, and progressive aggregation sampling strategy, our \textit{StableSR} serves as a strong baseline that inspires future research in adopting diffusion priors for restoration tasks.

\section{Related Work}

\noindent \textbf{Image Super-Resolution}.
Image Super-Resolution (SR) aims to restore an HR image from its degraded LR observation.
Early SR approaches \citep{dai2019second,dong2014learning,dong2015image,dong2016accelerating,he2019ode,xu2019towards,zhang2018image,chen2021pre,liang2021swinir,wang2018esrgan,ledig2017photo,sajjadi2017enhancenet,xu2017learning,zhou2020cross} assume a pre-defined degradation process, e.g., bicubic downsampling and blurring with known parameters.
While these methods can achieve appealing performance on the synthetic data with the same degradation, their performance deteriorates significantly in real-world scenarios due to the limited generalizability.

Recent works have moved their focus from synthetic settings to blind SR, where the degradation is unknown and similar to real-world scenarios.
Due to the lack of real-world paired data for training, some methods
\citep{fritsche2019frequency,maeda2020unpaired,wan2020bringing,wang2021unsupervised,wei2021unsupervised,zhang2021blind} propose to implicitly learn a degradation model from LR images in an unsupervised manner such as Cycle-GAN \citep{zhu2017unpaired} and contrastive learning \citep{oord2018representation}.
In addition to unsupervised learning, other approaches aim to explicitly synthesize LR-HR image pairs that resemble real-world data.
Specifically, BSRGAN \citep{zhang2021designing} and Real-ESRGAN \citep{wang2021realesrgan} present effective degradation pipelines for blind SR in real world.
Building upon such degradation pipelines, recent works based on diffusion models \citep{saharia2022image,sahak2023denoising} further show competitive performance on real-world image SR.
In this work, we consider an orthogonal direction of fine-tuning a diffusion model for SR.
In this way, the computational cost of network training could be reduced.
Moreover, our method allows the exploitation of generative prior encapsulated in the synthesis model, leading to better performance.

\noindent \textbf{Prior for Image Super-Resolution}.
To further enhance performance in complex real-world SR scenarios, numerous prior-based approaches have been proposed. These techniques deploy additional image priors to bolster the generation of faithful textures.
A straightforward method is reference-based SR \citep{zheng2018crossnet,zhang2019image,yang2020learning,jiang2021robust,zhou2020cross}. This involves using one or several reference high-resolution (HR) images, which share similar textures with the input low-resolution (LR) image, as an explicit prior to aid in generating the corresponding HR output. However, aligning features of the reference with the LR input can be challenging in real-world cases, and such explicit priors are not always readily available.
Recent works have moved away from relying on explicit priors, finding more promising performance with implicit priors instead.
Wang \etal~\citep{wang2018recovering} were the first to propose the use of semantic segmentation probability maps for guiding SR in the feature space. Subsequent works \citep{menon2020pulse,gu2020image,wang2021towards,pan2021exploiting,chan2021glean,chan2022glean,yang2021gan} employed pre-trained GANs by exploring the corresponding high-resolution latent space of the low-resolution input.
While effective, the implicit priors used in these approaches are often tailored for specific scenarios, such as limited categories \citep{wang2018recovering,gu2020image,pan2021exploiting,chan2021glean} and faces \citep{menon2020pulse,wang2021towards,yang2021gan}, and therefore lack generalizability for complex real-world SR tasks.
Other implicit priors for image SR include mixtures of degradation experts \citep{yu2018crafting,jie2022DASR} and VQGAN \citep{zhao2022rethinking,chen2022femasr,zhou2022codeformer}. However, these methods fall short, either due to insufficient prior expressiveness \citep{yu2018crafting,zhao2022rethinking,jie2022DASR} or inaccurate feature matching \citep{chen2022femasr}, resulting in output quality that remains less than satisfactory.

In contrast to existing strategies, we set our sights on exploring the robust and extensive generative prior found in pre-trained diffusion models \citep{nichol2022glide,rombach2021highresolution,ramesh2021zero,sahariaphotorealistic,ramesh2022hierarchical}.
While recent studies \citep{choi2021ilvr,avrahami2022blended,hulora,zhang2023adding,mou2023t2i} have highlighted the remarkable generative abilities of pre-trained diffusion models, the high-fidelity requirement inherent in super-resolution (SR) makes it unfeasible to directly adopt these methods for this task.
Our proposed StableSR, unlike LDM \citep{rombach2021highresolution}, does not necessitate training from scratch. Instead, it shares a similar idea to concurrent works~\citep{zhang2023adding,mou2023t2i} by fine-tuning directly on a frozen pre-trained diffusion model with only a small number of trainable parameters, leading to superior performance in a more efficient way.
In practice, our approach further shows comparable performance with follow-up works~\citep{lin2023diffbir,yu2024scaling}, which also exploit the diffusion prior but follow the ControlNet-like~\citep{zhang2023adding} framework. 
We provide a detailed comparison with these works in a following section.

\begin{figure*}[htbp]
\begin{center}
\centerline{\includegraphics[width=2.0\columnwidth]{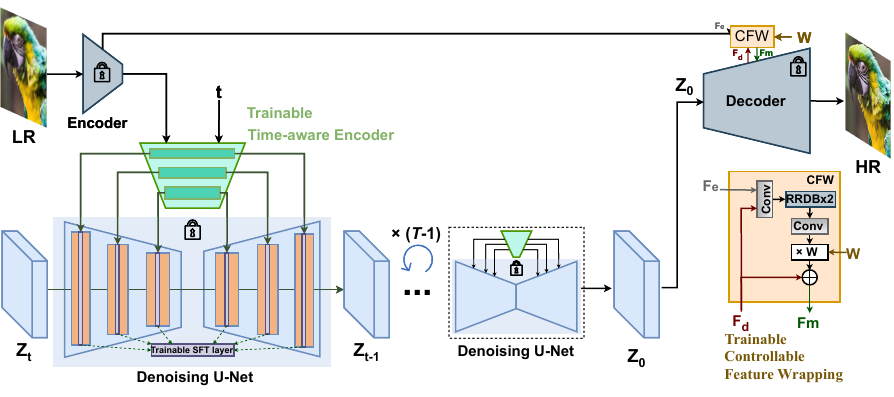}}
\caption{
Framework of StableSR.
We first finetune the time-aware encoder that is attached to a fixed pre-trained Stable Diffusion model.
Features are combined with trainable spatial feature transform (SFT) layers.
Such a simple yet effective design is capable of leveraging rich diffusion prior for image SR.
Then, the diffusion model is fixed. Inspired by CodeFormer~\citep{zhou2022codeformer}, we introduce a controllable feature wrapping (CFW) module to obtain a tuned feature $\bm{F}_m$ in a residual manner, given the additional information $\bm{F}_e$ from LR features and features $\bm{F}_d$ from the fixed decoder.
With an adjustable coefficient $w$, CFW can trade between quality and fidelity.
}
\label{network}
\end{center}
\vspace{-0.6cm}
\end{figure*}

\vspace{-0.2cm}
\section{Methodology}
Our method employs diffusion prior for SR. Inspired by the generative capabilities of Stable Diffusion~\citep{rombach2021highresolution}, we use it as the diffusion prior in our work, hence the name \textit{StableSR} for our method.
The main component of StableSR is a time-aware encoder, which is trained along with a frozen Stable Diffusion model to allow for conditioning based on the input image.
To further facilitate a trade-off between realism and fidelity, depending on user preference, we follow CodeFormer~\citep{zhou2022codeformer} to introduce an optional controllable feature wrapping module. The overall framework of StableSR is depicted in Fig.~\ref{network}.

\vspace{-0.2cm}
\subsection{Guided Finetuning with Time Awareness}
\label{sec:enc}
To exploit the prior knowledge of Stable Diffusion for SR, we establish the following constraints when designing our model:
1) The resulting model must have the ability to generate a plausible HR image, conditioned on the observed LR input. This is vital because the LR image is the only source of structural information, which is crucial for maintaining high fidelity.
2) The model should introduce only minimal alterations to the original Stable Diffusion model to prevent disrupting the prior encapsulated within it.

\noindent\textbf{Feature Modulation}.
While several existing approaches \citep{nichol2022glide,rombach2021highresolution,hertz2022prompt,feng2022training,balaji2022eDiff-I} have successfully controlled the generated semantic structure of a diffusion model via cross-attention, such a strategy can hardly provide detailed and high-frequency guidance due to insufficient inductive bias \citep{liu2021swin}.
To more accurately guide the generation process, we adopt an additional encoder to extract multi-scale features $\{\bm{F}^n\}^{N}_{n=1}$ from the degraded LR image features, and use them to modulate the intermediate feature maps $\{\bm{F}^n_{\rm dif}\}^{N}_{n=1}$ of the residual blocks in Stable Diffusion via spatial feature transformations (SFT) \citep{wang2018recovering}:
\begin{equation}
\hat{\bm{F}}^n_{\rm dif} = (1 + \bm{\alpha}^n) \odot \bm{F}^n_{\rm dif} + \bm{\beta}^n; ~
\bm{\alpha}^n, \bm{\beta}^n = \mathcal{M}^n_{\theta}(\bm{F}^n),
\label{eq:sft}
\end{equation}
where $\bm{\alpha}^n$ and $\bm{\beta}^n$ denote the affine parameters in SFT and $\mathcal{M}^n_{\theta}$ denotes a small network consisting of several convolutional layers.
Here $n$ indices the spatial scale of the UNet~\citep{ronneberger2015u} architecture in Stable Diffusion.

During finetuning, we freeze the weights of Stable Diffusion and train only the encoder and SFT layers.
This strategy allows us to insert structural information extracted from the LR image without destroying the generative prior captured by Stable Diffusion.

\begin{figure*}[htbp]
	\begin{center}		\centerline{\includegraphics[width=2.1\columnwidth]{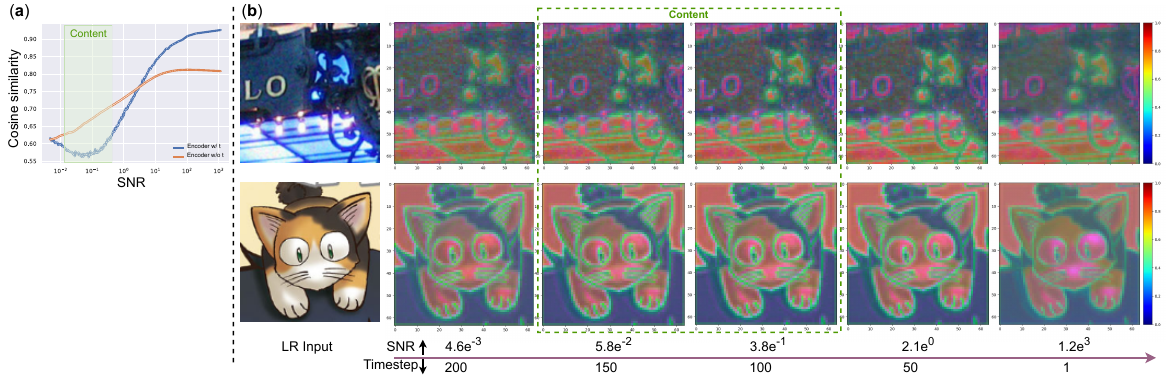}}
    \vspace{-0.2cm}
		\caption{
        In contrast to a conditional encoder without time embedding, the one equipped with time embedding can adaptively supply guidance to the pre-trained diffusion models.
        (a), we gauge the cosine similarity between the diffusion model's features pre- and post-SFT at various timesteps, which echoes the strength of the condition originating from the encoder.
        (b), we further visualize the features of the conditional encoder extracted from the LR image. As shown, the encoder is inclined to provide sharp features when the SNR hovers around $5\text{e}^{-2}$. This is precisely when the diffusion model requires substantial guidance to generate the desired high-resolution image content.
        Interestingly, this observation aligns with the findings in \citep{choi2022perception}.
        }
	\label{fea_cos}
	\end{center}
 \vspace{-0.6cm}
\end{figure*}

\noindent\textbf{Time-aware Guidance}.
We find that incorporating temporal information through a time-embedding layer in our encoder considerably enhances both the quality of generation and the fidelity to the ground truth, since it can adaptively adjust the condition strength derived from the LR features.
Here, we analyze this phenomenon from a signal-to-noise ratio (SNR) standpoint and later quantitatively and qualitatively validate it in the ablation study.

During the generation process, the SNR of the produced image progressively increases as noise is incrementally removed. A recent study~\citep{choi2022perception} indicates that image content is rapidly populated when the SNR approaches $5\text{e}^{-2}$.
In line with this observation,
we notice that the time embedding enables the conditional encoder to provide stronger guidance within
the range where the signal-to-noise ratio (SNR) hits $5\text{e}^{-2}$.
This is essential because the content generated at this stage significantly influences the super-resolution performance of our method.
To further substantiate this, since the conditional features are inserted into the diffusion prior via SFT layers, we employ the cosine similarity between the features of Stable Diffusion before and after the SFT to measure the condition strength provided by the encoder.
The cosine similarity values at different timesteps are plotted in Fig.~\ref{fea_cos}-(a). As can be observed, the cosine similarity reaches its minimum value around an SNR of $5\text{e}^{-2}$, indicative of the strongest conditions imposed by the encoder.
In addition, we also depict the feature maps extracted from our specially designed encoder in Fig.~\ref{fea_cos}-(b). It is noticeable that the features around the SNR point of $5\text{e}^{-2}$ are sharper and contain more detailed image structures.
We hypothesize that these adaptive feature conditions can furnish more comprehensive guidance for SR.

\noindent\textbf{Color Correction}.
Diffusion models can occasionally exhibit color shifts, as noted in \citep{choi2022perception}. To address this issue, we perform color normalization on the generated image to align its mean and variance with those of the LR input.
In particular, if we let $\bm{x}$ denote the LR input and $\hat{\bm{y}}$ represent the generated HR image, the color-corrected output, $\bm{y}$, is calculated as follows:
\begin{equation}
     \bm{y}^c = \frac{\hat{\bm{y}}^c-\bm{\mu}_{\hat{\bm{y}}}^{c}}{\bm{\sigma}_{\hat{\bm{y}}}^c} \cdot \bm{\sigma}_x^c + \bm{\mu}_x^c,
     \label{eq:color_chn_matching}
\end{equation}
where $c\in \{r, g, b\}$ denotes the color channel, $\bm{\mu}^c_{\hat{\bm{y}}}$ and $\bm{\sigma}^c_{\hat{\bm{y}}}$ (or $\bm{\mu}^c_x$ and $\bm{\sigma}^c_x$) are the mean and standard variance estimated from the $c$-th channel of $\hat{\bm{y}}$ (or $\bm{x}$), respectively.
We find that this simple correction suffices to remedy the color difference.

Though pixel color correction via channel matching can improve color fidelity, we notice that it may suffer from limited color correction ability due to the lack of pixel-wise controllability.
The main reason is that it only introduces global statistics, i.e., channel-wise mean and variance of the input for color correction, ignoring pixel-wise semantics.
Besides adopting color correction in the pixel domain, we further propose wavelet-based color correction for better visual performance in some cases.
Wavelet color correction directly introduces the low-frequency part from the input since the color information belongs to the low-frequency components, while the degradations are mostly high-frequency components.
In this way, we can improve the color fidelity of the results without perceptibly affecting the generated quality.
Given any image $\bm{I}$, we extract its high-frequency component $\bm{H}^i$ and low-frequency component $\bm{L}^i$ at the $i$-th ($1\le i\le l$) scale via the wavelet decomposition, i.e.,
\begin{equation}
    \bm{L}^i = \mathcal{C}_i (\bm{L}^{i-1}, \bm{k}), ~ \bm{H}^i = \bm{L}^{i-1} - \bm{L}^i,
\end{equation}
where $\bm{L}^0 = \bm{I}$, $\mathcal{C}_i$ denotes the convolutional operator with a dilation of $2^i$, and $k$ is the convolutional kernel defined as:
\begin{equation}
    \bm{k} = \begin{bmatrix}
        \sfrac{1}{16} & \sfrac{1}{8} & \sfrac{1}{16} \\
        \sfrac{1}{8}  & \sfrac{1}{4} & \sfrac{1}{8} \\
        \sfrac{1}{16} & \sfrac{1}{8} & \sfrac{1}{16}
    \end{bmatrix}.
\end{equation}
By denoting the $l$-th low-frequency and high-frequency components of $\bm{x}$ (or $\hat{\bm{y}}$) as $\bm{L}^l_x$ and $\bm{H}^l_x$ (or $\bm{L}^l_y$ and $\bm{H}^l_y$), the desired HR output $\bm{y}$ is formulated as follows:
\begin{equation}
    \bm{y} = \bm{H}^l_y + \bm{L}^l_x.
    \label{eq:color_wavlet}
\end{equation}
Intuitively, we replace the low-frequency component $\bm{L}^l_y$ of $\hat{\bm{y}}$ with $\bm{L}^l_x$ to correct the color bias.
By default, we adopt color correction in the pixel domain for simplicity.

\subsection{Fidelity-Realism Trade-off}
\label{sec:imp}
Although the output of the proposed approach is visually compelling, it often deviates from the ground truth due to the inherent stochasticity of the diffusion model. Drawing inspiration from CodeFormer \citep{zhou2022codeformer}, we introduce a Controllable Feature Wrapping (CFW) module to flexibly manage the balance between realism and fidelity.
Unlike CodeFormer, there are multiple sampling steps for generating a sample during inference and we cannot finetune the CFW module directly. To overcome this problem, we first generate synthetic LR-HR pairs following the same degradation pipeline with the diffusion training stage. Then, the latent codes $\bm{Z}_0$ can be obtained using our finetuned diffusion model given the LR images as conditions. Finally, CFW can be trained using the generated data.

Since Stable Diffusion is implemented in the latent space of an autoencoder, it is natural to leverage the encoder features of the autoencoder to modulate the corresponding decoder features for further fidelity improvement.
Let $\bm{F}_e$ and $\bm{F}_d$ be the encoder and decoder features, respectively.
We introduce an adjustable coefficient $w \in [0,1]$ to control the extent of modulation:
\begin{equation}
    \bm{F}_m = \bm{F}_d + \mathcal{C}(\bm{F}_e, \bm{F}_d;\bm{\theta}) \times w,
\end{equation}
where $\mathcal{C}(\cdot;\bm{\theta})$ represents convolution layers with trainable parameter $\bm{\theta}$.
The overall framework is shown in Fig.~\ref{network}.

In this design, a small $w$  exploits the generation capability of Stable Diffusion, leading to outputs with high realism under severe degradations.
In contrast, a large $w$ allows stronger structural guidance from the LR image, enhancing fidelity.
We observe that $w\,{=}\,0.5$ achieves a good balance between quality and fidelity.
%
Note that we only train CFW in this particular stage.
In practice, we notice that CFW involves additional GPU memory and the improvement can be subtle in some cases.
Thus, we make it optional for different real-world applications.

\begin{figure}[t]
\begin{center}
\centerline{\includegraphics[width=1.0\columnwidth]{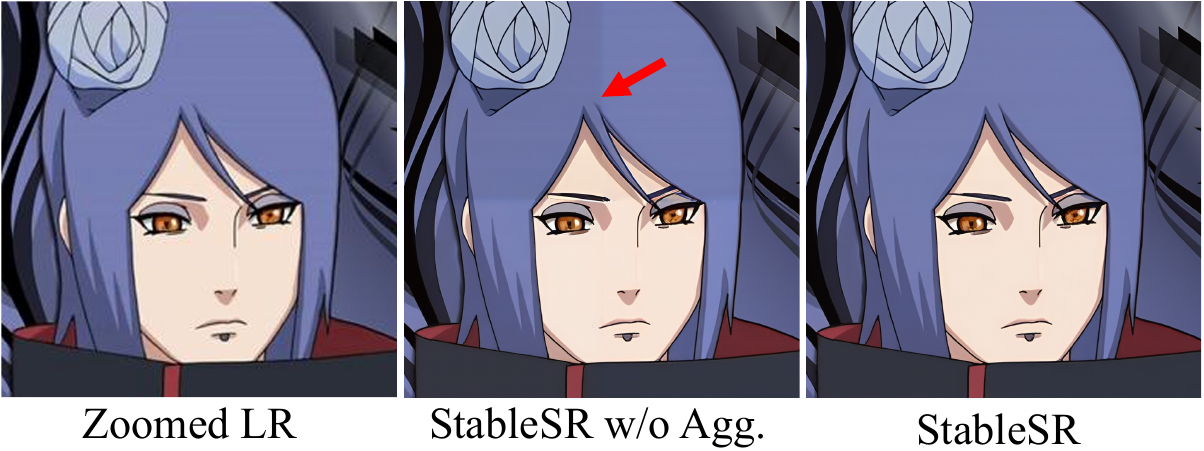}}
\caption{When dealing with images beyond $512 \times 512$, StableSR (w/o aggregation sampling) suffers from obvious block inconsistency by chopping the image into several tiles, processing them separately, and stitching them together.
With our proposed aggregation sampling, StableSR can achieve consistent results on large images.
The resolution of the shown figure is $1024 \times 1024$.
}
\label{abla_size}
\end{center}
\vspace{-0.6cm}
\end{figure}

\subsection{Aggregation Sampling}
\label{sec:cfw}
Due to the heightened sensitivity of the attention layers in Stable Diffusion with respect to the image resolution, it tends to produce inferior outputs for resolutions differing from its training settings, specifically $512{\times}512$. This, in effect, constrains the practicality of StableSR.

A common workaround involves splitting the larger image into several overlapping smaller patches and processing each individually. While this strategy often yields good results for conventional CNN-based SR methods, it is not directly applicable to the diffusion paradigm. This is because discrepancies between patches are compounded and magnified over the course of diffusion iterations. A typical failure case is illustrated in Fig.~\ref{abla_size}.

Inspired by Jim\'{e}nez~\citep{jimenez2023mixture}, we apply a progressive patch aggregation sampling algorithm to handle images of arbitrary resolutions.
Specifically, we begin by encoding the LR image into a latent feature map $\bm{F}\in \mathcal{R}^{h\times w}$, which is then subdivided into $M$ overlapping small patches $\{\bm{F}_{\Omega_n}\}_{n=1}^{M}$, each with a resolution of $64\times64$ - matching the training resolution\footnote{The downsampling scale factor of the autoencoder in Stable Diffusion is $8\times$.}.
Here, $\Omega_n$ is the coordinate set of the $n$th patch in $\bm{F}$.
During each timestep in the reverse sampling, each patch is individually processed through StableSR, with the processed patches subsequently aggregated.
To integrate overlapping patches, a weight map $\bm{w}_{\Omega_n}\in \mathcal{R}^{h\times w}$ whose entries follow up a Gaussian filter in $\Omega_n$ and 0 elsewhere is generated for each patch $\bm{F}_{\Omega_n}$.
Overlapping pixels are then weighted in accordance with their respective Gaussian weight maps.
In particular, we follow Jim\'{e}nez~\citep{jimenez2023mixture} to define a padding function $f(\cdot)$ that expands any patch of size $64\times64$ to the resolution of $h\times w$ by filling zeros outside the region $\Omega_n$.
This procedure is reiterated until reaching the final iteration.

Given the output of each patch as $\epsilon_{\bm{\theta}}(\bm{Z}^{(t)}_{\Omega_n}, \bm{F}_{\Omega_n},t)$, where $\bm{Z}^{(t)}_{\Omega_n}$ is the $n$th patch of the noisy input $\bm{Z}^{(t)}$ and $\bm{\theta}$ is the parameters of the diffusion model,
the results of all the patches aggregated together can be formulated as follows:
\begin{equation}
       \epsilon_{\bm{\theta}}(\bm{Z}^{(t)}, \bm{F}, t) = \sum_{n=1}^M \frac{\bm{w}_{\Omega_n}}{\hat{\bm{w}}} \odot f\left( \epsilon_{\bm{\theta}}\left(\bm{Z}^{(t)}_{\Omega_n}, \bm{F}_{\Omega_n}, t\right) \right),
       \label{eq:noise_mix}
\end{equation}
where $\hat{\bm{w}}=\sum_n \bm{w}_{\Omega_n}$.
Based on $\epsilon_{\bm{\theta}}(\bm{Z}^{(t)}, \bm{F}, t)$, we can obtain $\bm{Z}^{(t-1)}$ according to the sampling procedure, denoted as ${\rm Sampler}(\bm{Z}^{(t)}, \epsilon_{\bm{\theta}}(\bm{Z}^{(t)}, \bm{F},t))$, in the diffusion model.
Subsequently, we re-split $\bm{Z}^{(t-1)}$ into over-lapped patches and repeat the above steps until $t=1$.
The whole process is summed up in Algorithm \ref{alg:mos}.
Our experiments suggest that this progressive aggregation method substantially mitigates discrepancies in the overlapped regions, as depicted in Fig.~\ref{abla_size}.
More details can be found in the supplementary material.

\begin{algorithm}
\caption{Progressive Patch Aggregation}\label{alg:mos}
\begin{algorithmic}[1]
\Require Cropped Regions $\{\Omega_n\}_{n=1}^M$, diffusion steps $T$, LR latent features $\bm{F}$.
\State Initialize $\bm{w}_{\Omega_n}$ and $\hat{\bm{w}}$  
\State $\bm{Z}^{(T)} \sim {\cal N}(0, {\mathbb I})$  
\For{$t \in [T, \ldots, 0]$}  
    \For{$n \in [1, \ldots, M]$}
        \State Compute $\epsilon_{\bm{\theta}}\left(\bm{Z}^{(t)}_{\Omega_n}, \bm{F}_{\Omega_n}, t\right)$
    \EndFor
    \State Compute $\epsilon_{\bm{\theta}}(\bm{Z}^{(t)}, \bm{F}, t)$ following Eq.~\eqref{eq:noise_mix}
    \State $\bm{Z}^{(t-1)}={\rm Sampler}(\bm{Z}^{(t)}, \epsilon_{\bm{\theta}}(\bm{Z}^{(t)}, \bm{F}, t))$ 
\EndFor
\State \Return $\bm{Z}_0$
\end{algorithmic}
\end{algorithm}
\vspace{-1.5cm}

\section{Experiments}
\label{sec_exp}
\subsection{Implementation Details}
StableSR is built based on Stable Diffusion 2.1-base\footnote{https://huggingface.co/stabilityai/stable-diffusion-2-1-base}.
Our time-aware encoder is similar to the contracting path of the denoising U-Net in Stable Diffusion but is much more lightweight (${\sim}$105M, including SFT layers).
SFT layers are inserted in each residual block of Stable Diffusion for effective control.
We finetune the diffusion model of StableSR for $117$ epochs with a batch size of $192$, and the prompt is fixed as null.
We follow Stable Diffusion to use Adam \citep{kingma2014adam} optimizer and the learning rate is set to $5 \times 10^{-5}$.
The training process is conducted on $512 \times 512$ resolution with 8 NVIDIA Tesla 32G-V100 GPUs.
For inference, we adopt DDPM sampling \citep{ho2020denoising} with 200 timesteps.
To handle images with arbitrary sizes, we adopt the proposed aggregation sampling strategy for images beyond $512 \times 512$.
As for images under $512 \times 512$, we first enlarge the LR images such that the shorter side has a length of $512$ and rescale the results back to target resolutions after generation.

To train CFW, we first generate 100k synthetic LR-HR pairs with $512 \times 512$ resolution following the degradation pipeline in Real-ESRGAN~\citep{wang2021realesrgan}.
Then, we adopt the finetuned diffusion model to generate the corresponding latent codes $\bm{Z}_0$ given the above LR images as conditions.
The training losses are almost the same as the autoencoder used in LDM \citep{rombach2021highresolution}, except that we use a fixed adversarial loss weight of $0.025$ rather than a self-adjustable one.

\begin{table*}[htbp]
 \newcommand{\tabincell}[2]{\begin{tabular}{@{}#1@{}}#2\end{tabular}}
 \begin{center}
  \caption{Quantitative comparison with state-of-the-art methods on both synthetic and real-world benchmarks. \rf{Red} and \bd{blue} colors represent the best and second best performance, respectively.}
  \label{metric_tab}
\resizebox{\textwidth}{!}{
  \begin{tabular}{c|c|ccccccccc}
                \hline Datasets & Metrics & \makecell[c]{RealSR} & \makecell[c]{BSRGAN} & \makecell[c]{DASR} & \makecell[c]{Real-ESRGAN+} & \makecell[c]{FeMaSR} & \makecell[c]{LDM} & \makecell[c]{SwinIR-GAN} & \makecell[c]{IF\_III} & \makecell[c]{\textbf{StableSR}} \\
                \hline \multirow{6}{*}{\makecell[c]{DIV2K Valid}} & PSNR $\uparrow$ & \rf{24.62} & \bd{24.58} & 24.47 & 24.29 & 23.06 & 23.32 & 23.93 & 23.36 & 23.26 \\
                 & SSIM $\uparrow$ & 0.5970 & 0.6269 & 0.6304 & \rf{0.6372} & 0.5887 & 0.5762 & \bd{0.6285} & 0.5636 & 0.5726 \\
                 & LPIPS $\downarrow$ & 0.5276 & 0.3351 & 0.3543 & \rf{0.3112} & 0.3126 & 0.3199 & 0.3160 & 0.4641 & \bd{0.3114} \\
                & FID $\downarrow$ & 49.49 & 44.22 & 49.16 & 37.64 & 35.87 & \bd{26.47} & 36.34 & 37.54 & \rf{24.44} \\
                & CLIP-IQA $\uparrow$ & 0.3534 & 0.5246 & 0.5036 & 0.5276 & 0.5998 & \bd{0.6245} & 0.5338 & 0.3980 & \rf{0.6771}\\
                & MUSIQ $\uparrow$ & 28.57 & 61.19 & 55.19 & 61.05 & 60.83 & \bd{62.27} & 60.22 & 43.71 & \rf{65.92} \\
                \hline
                \hline \multirow{5}{*}{\makecell[c]{RealSR}}
                & PSNR $\uparrow$ & \rf{27.30} & 26.38 & \bd{27.02} & 25.69 & 25.06 & 25.46 & 26.31 & 25.47 & 24.65 \\
                 & SSIM $\uparrow$ & 0.7579 & 0.7651 & \bd{0.7707} & 0.7614 & 0.7356 & 0.7145 & \rf{0.7729} & 0.7067 & 0.7080 \\
                 & LPIPS $\downarrow$ & 0.3570 & \bd{0.2656} & 0.3134 & 0.2709 & 0.2937 & 0.3159 & \rf{0.2539} & 0.3462 & 0.3002 \\
                 & CLIP-IQA $\uparrow$ & 0.3687 & 0.5114 & 0.3198 & 0.4495 & 0.5406 & \bd{0.5688} &  0.4360 & 0.3482 & \rf{0.6234}\\
                & MUSIQ $\uparrow$ & 38.26 & \bd{63.28} & 41.21 & 60.36 & 59.06 & 58.90 & 58.70 & 41.71 & \rf{65.88}\\
                \hline \multirow{5}{*}{\makecell[c]{DRealSR}}
                & PSNR $\uparrow$ & \rf{30.19} & 28.70 & \bd{29.75} & 28.62 & 26.87 & 27.88 & 28.50 & 28.66 & 28.03 \\
                 & SSIM $\uparrow$ & \bd{0.8148} & 0.8028 & \rf{0.8262} & 0.8052 & 0.7569 & 0.7448 & 0.8043 & 0.7860 & 0.7536 \\
                 & LPIPS $\downarrow$ & 0.3938 & 0.2858 & 0.3099 & \bd{0.2818} & 0.3157 & 0.3379 & \rf{0.2743} & 0.3853 & 0.3284 \\
                & CLIP-IQA $\uparrow$ & 0.3744 & 0.5091 & 0.3813 & 0.4515 & 0.5634 &  \bd{0.5756} & 0.4447 & 0.2925 & \rf{0.6357}\\
                & MUSIQ $\uparrow$ & 26.93 & \bd{57.16} & 42.41 & 54.26 & 53.71 & 53.72 & 52.74 & 30.71 & \rf{58.51} \\
                \hline \multirow{2}{*}{\makecell[c]{DPED-iphone}} & CLIP-IQA $\uparrow$ & 0.4496 & 0.4021 &  0.2826 & 0.3389 & \rf{0.5306} & 0.4482 & 0.3373 & 0.2962 & \bd{0.4799}\\
                & MUSIQ $\uparrow$ & 45.60 & 45.89 & 32.68 & 42.42 & \bd{49.95} & 44.23 & 43.30 & 37.49 & \rf{50.48}\\
                \hline
  \end{tabular}
}
 \end{center}
 \vspace{-0.25cm}
\end{table*}

\subsection{Experimental Settings}
\noindent\textbf{Training Datasets.}
We adopt the degradation pipeline of Real-ESRGAN \citep{wang2021realesrgan} to synthesize LR/HR pairs on DIV2K \citep{agustsson2017ntire}, DIV8K \citep{gu2019div8k}, Flickr2K \citep{timofte2017ntire} and OutdoorSceneTraining \citep{wang2018recovering} datasets.
We additionally add 5000 face images from the FFHQ dataset
 \citep{karras2019style} for general cases.

\noindent\textbf{Testing Datasets.}
We evaluate our approach on both synthetic and real-world datasets.
For synthetic data, we follow the degradation pipeline of Real-ESRGAN \citep{wang2021realesrgan} and generate 3k LR-HR pairs from DIV2K validation set \citep{agustsson2017ntire}.
The resolution of LR is $128 \times 128$ and that of the corresponding HR is $512 \times 512$.
Note that for StableSR, the inputs are first upsampled to the same size as the outputs before inference.
For real-world datasets, we follow common settings to conduct comparisons on RealSR \citep{cai2019toward}, DRealSR \citep{wei2020component} and DPED-iPhone \citep{ignatov2017dslr}.
We further collect 40 images from the Internet for comparison.

\begin{figure*}[!h]
\begin{center}
\centerline{\includegraphics[width=2.05\columnwidth]{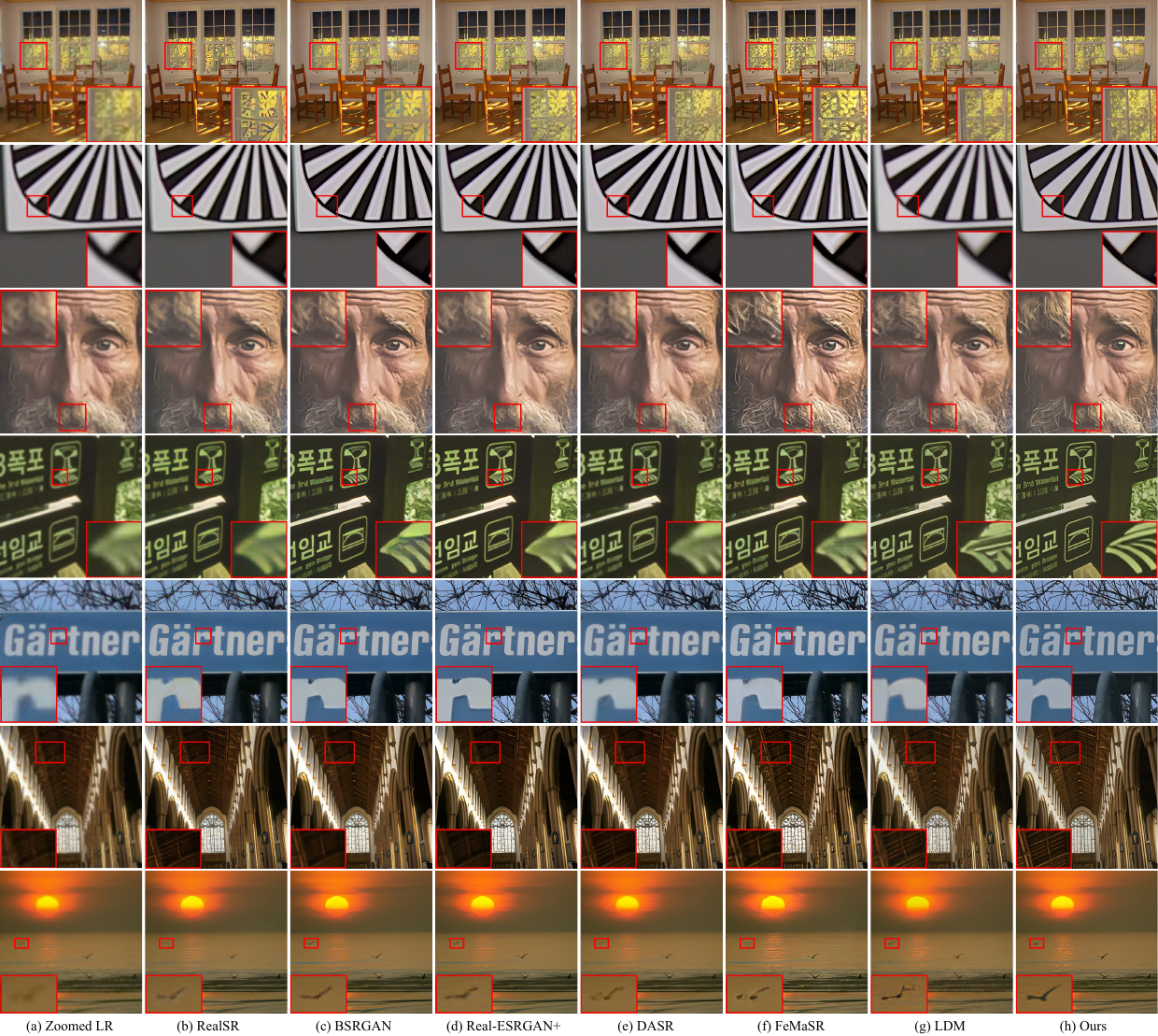}}
\caption{Qualitative comparisons on several representative real-world samples ($128 \rightarrow 512$). Our StableSR is capable of removing artifacts and generating realistic details. (\textbf{Zoom in for details})}
\label{vis_fig_real}
\end{center}
\vspace{-0.7cm}
\end{figure*}

\noindent\textbf{Compared Methods.}
To verify the effectiveness of our approach, we compare our StableSR with several state-of-the-art methods\footnote{SR3 \citep{saharia2022image} is not included since its official code is unavailable.}, \ie, RealSR\footnote{We use the latest official model DF2K-JPEG.} \citep{ji2020real}, BSRGAN \citep{zhang2021designing}, Real-ESRGAN+ \citep{wang2021realesrgan}, DASR \citep{jie2022DASR}, FeMaSR \citep{chen2022femasr}, latent diffusion model (LDM) \citep{rombach2021highresolution}, SwinIR-GAN\footnote{We use the latest official SwinIR-GAN model, \ie, 003\_realSR\_BSRGAN\_DFOWMFC\_s64w8\_SwinIR-L\_x4\_GAN.pth.} \citep{liang2021swinir}, and DeepFloyd IF\_III \citep{deepfloyd2023}.
Since LDM is officially trained on images with $256 \times 256$ resolution, we finetune it following the same training settings of StableSR for a fair comparison.
For other methods, we directly use the official code and models for testing.
Note that the results in this section are obtained on the same resolution with training, \ie, $128 \times 128$.
Specifically, for images from \citep{cai2019toward,wei2020component,ignatov2017dslr}, we crop them at the center to obtain patches with $128 \times 128$ resolution.
For other real-world images, we first resize them such that the shorter sides are $128$ and then apply center cropping.
As for other resolutions, one example of StableSR on real-world images under $1024 \times 1024$ resolution is shown in Fig.~\ref{abla_size}.
More results are provided in the supplementary material.

\noindent \textbf{Evaluation Metrics.}
For benchmarks with paired data, \ie, DIV2K Valid, RealSR and DRealSR, we employ various perceptual metrics including LPIPS\footnote{We use LPIPS-ALEX by default.}\citep{zhang2018perceptual}, FID \citep{heusel2017gans}, CLIP-IQA \citep{wang2022exploring} and MUSIQ \citep{ke2021musiq} to evaluate the perceptual quality of generated images.
PSNR and SSIM scores (evaluated on the luminance channel in YCbCr color space) are also reported for reference.
Since ground-truth images are unavailable in DPED-iPhone \citep{ignatov2017dslr}, we follow existing methods \citep{wang2021realesrgan,chen2022femasr} to report results on no-reference metrics \ie, CLIP-IQA and MUSIQ for perceptual quality evaluation.
Besides, we further conduct a user study on $16$ real-world images to verify the effectiveness of our approach against existing methods.

\subsection{Comparison with Existing Methods}
\noindent\textbf{Quantitative Comparisons.}
We first show the quantitative comparison on the synthetic DIV2K validation set and three real-world benchmarks.
As shown in Table \ref{metric_tab}, our approach outperforms state-of-the-art SR methods in terms of multiple perceptual metrics, including FID, CLIP-IQA and MUSIQ.
Specifically, on synthetic benchmark DIV2K Valid, our StableSR ($w = 0.5$) achieves a $24.44$ FID score, which is $7.7\%$ lower than LDM and at least $32.9 \%$ lower than other GAN-based methods.
Besides, our StableSR ($w = 0.5$) achieves the highest CLIP-IQA scores on the two commonly used real-world benchmarks \citep{cai2019toward,wei2020component}, suggesting the superiority of StableSR.
While we notice that StableSR achieves inferior performance on metrics including PSNR, SSIM and LPIPS compared with non-diffusion methods, these metrics only reflect certain aspects of performance~\citep{ledig2017photo,wang2018esrgan,blau2018perception}.
Besides, the previous non-diffusion methods tend to directly use $\ell_2$ losses and perceptual loss between the predictions and the corresponding ground truths for training, which are closely related to the calculation of PSNR and LPIPS, respectively.
Different from previous methods, diffusion models~\citep{ho2020denoising,rombach2021highresolution} only apply $\ell_2$ loss between the predicted and the ground-truth noise.
We conjecture this is an important factor that makes diffusion models less competitive on these metrics, as observed by the recent work~\citep{yue2022difface}.
Moreover, previous methods usually fail to restore faithful textures and generate blurry results, as shown in Fig.~\ref{vis_fig_real}.
In contrast, our StableSR is capable of generating sharp images with realistic details.

\begin{figure*}[htbp]
	\begin{center}
		\centerline{\includegraphics[width=2.05\columnwidth]{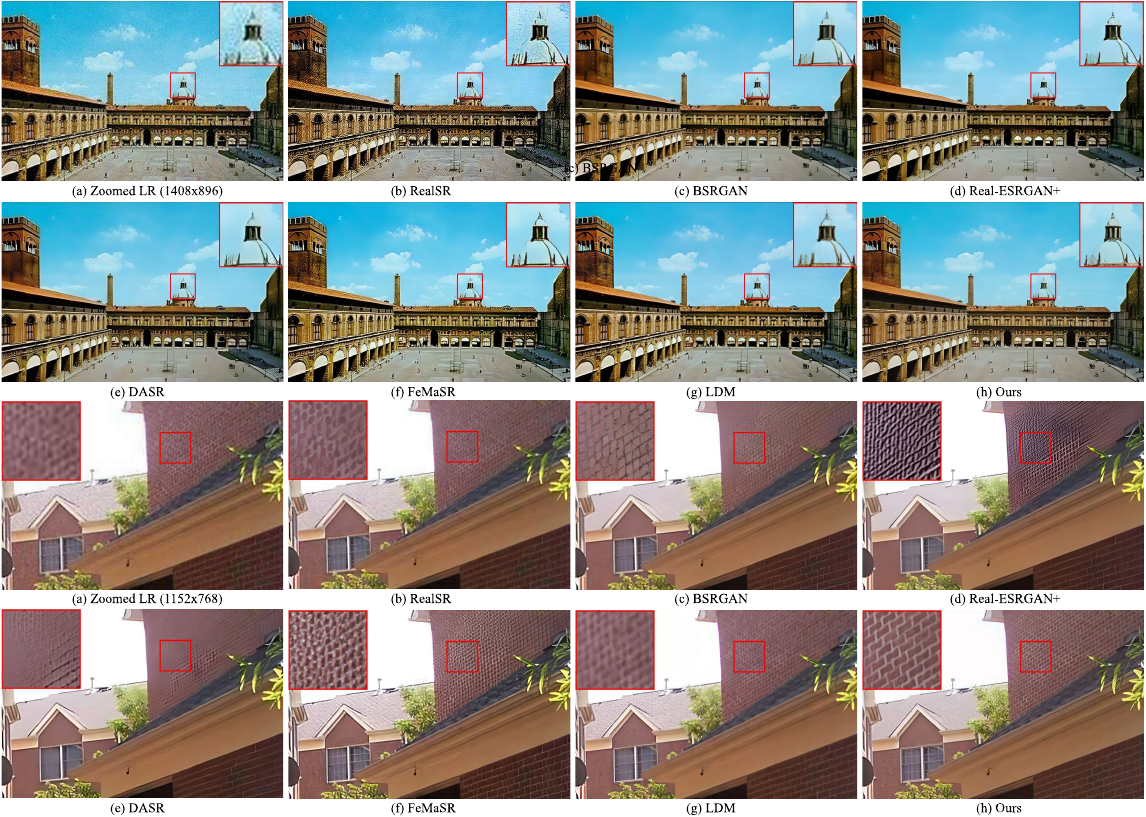}}
		\caption{Qualitative comparisons on real-world images with diverse resolutions beyond $512 \times 512$. Our StableSR still outperforms other methods with more vivid details and less annoying artifacts. (\textbf{Zoom in for details})}
	\label{fig_arb}
    \vspace{-0.5cm}
	\end{center}
\end{figure*}

\noindent\textbf{Qualitative Comparisons.}
To demonstrate the effectiveness of our method, we present visual results on real-world images from both real-world benchmarks \citep{cai2019toward,wei2020component} and the internet in Fig.~\ref{vis_fig_real} and Fig.~\ref{fig_arb}.
It is observed that StableSR outperforms previous methods in both artifact removal and detail generation.
Specifically, StableSR is able to generate faithful details, as shown in the first row of Fig.~\ref{vis_fig_real}, while other methods either show blurry results (DASR, BSRGAN, Real-ESRGAN+, LDM) or unnatural details (RealSR, FeMaSR).
Moreover, as shown in the fourth row of Fig.~\ref{vis_fig_real}, StableSR generates sharp edges without obvious degradations, whereas other state-of-the-art methods generate blurry results.
Figure~\ref{fig_arb} further demonstrates the superiority of StableSR on images beyond $512 \times 512$.

\begin{figure}[tbp]
\begin{center}
\centerline{\includegraphics[width=1.0\columnwidth]{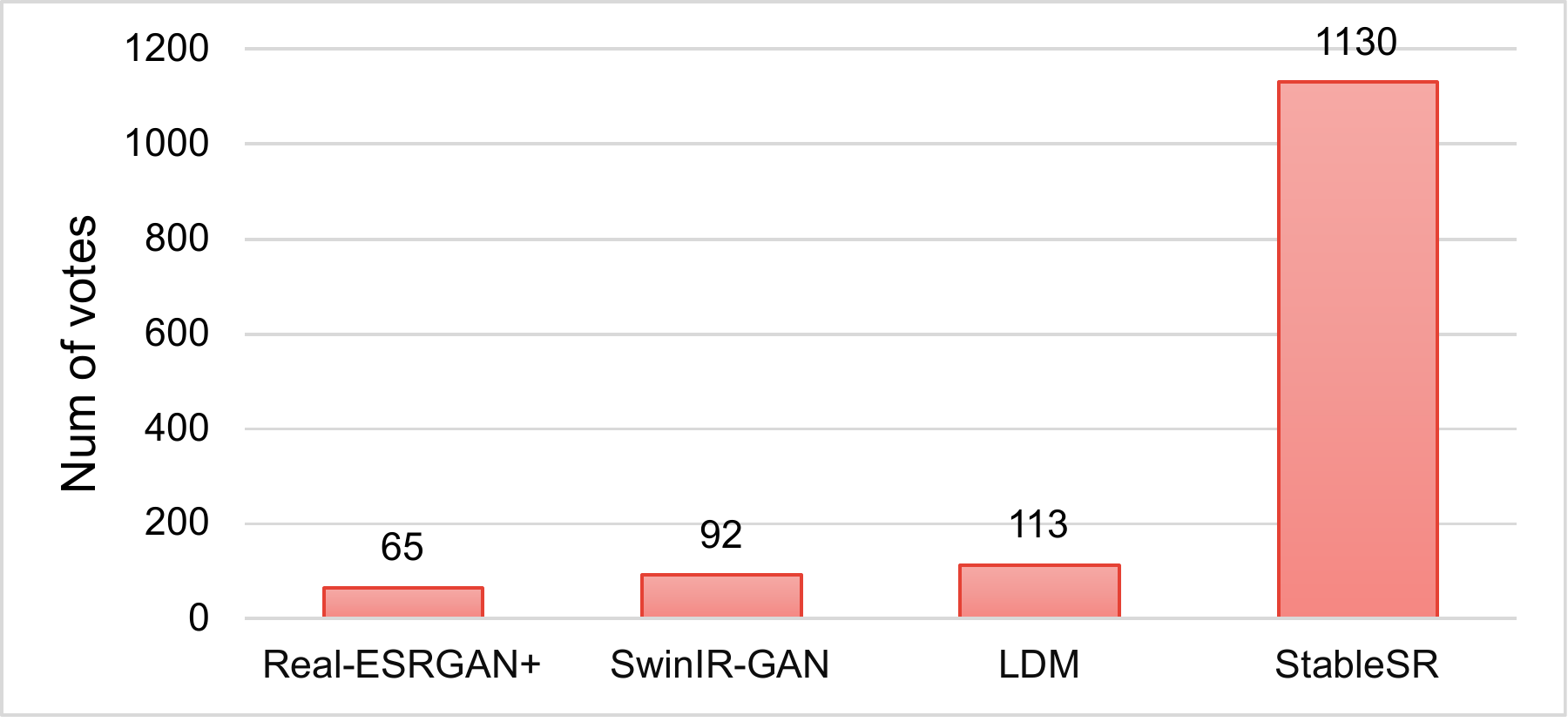}}
\caption{User study on 40 real-world images evaluated by 35 subjects. Given one LR image, the subjects are asked to choose the best HR image generated from the methods including StableSR,  LDM, Real-ESRGAN+ and SwinIR w/ GAN.
The large number of votes gained by StableSR indicates its potential capability for real-world SR applications.}
\label{fig_mos}
\end{center}
\vspace{-0.7cm}
\end{figure}

\noindent\textbf{User Study.}
To further examine the effectiveness of StableSR, we conduct a user study on 40 real-world LR images collected from the Internet.
To alleviate potential bias, the collected real-world images contain diverse content, e.g., natural images with and without objects, and photos with texts and faces.
The order of the images as well as the options are also randomly shuffled.
We further provide the link\footnote{https://forms.gle/gsLyVr6pSkAEbW8J9} of our user study for reference.
We compare our approach with three commonly used SR methods with competitive performance, \ie, Real-ESRGAN+, SwinIR-GAN and LDM.
Given a LR image as reference, the subject is asked to choose the best HR image generated from the four methods, i.e.,  StableSR, Real-ESRGAN+, SwinIR-GAN and LDM.
Given the 40 LR images with the three compared methods, there are 35 subjects for evaluation, resulting in  $40 \times 35 = 1400$ votes in total.
As depicted in Fig.~\ref{fig_mos}, by gaining over 80\% of the votes, StableSR shows its potential capability for real-world SR applications.
However, we also notice that StableSR may struggle in dealing with small texts, faces and patterns, indicating there is still room for improvement.

%

\begin{figure}[!t]
\begin{center}
\centerline{\includegraphics[width=1.0\columnwidth]{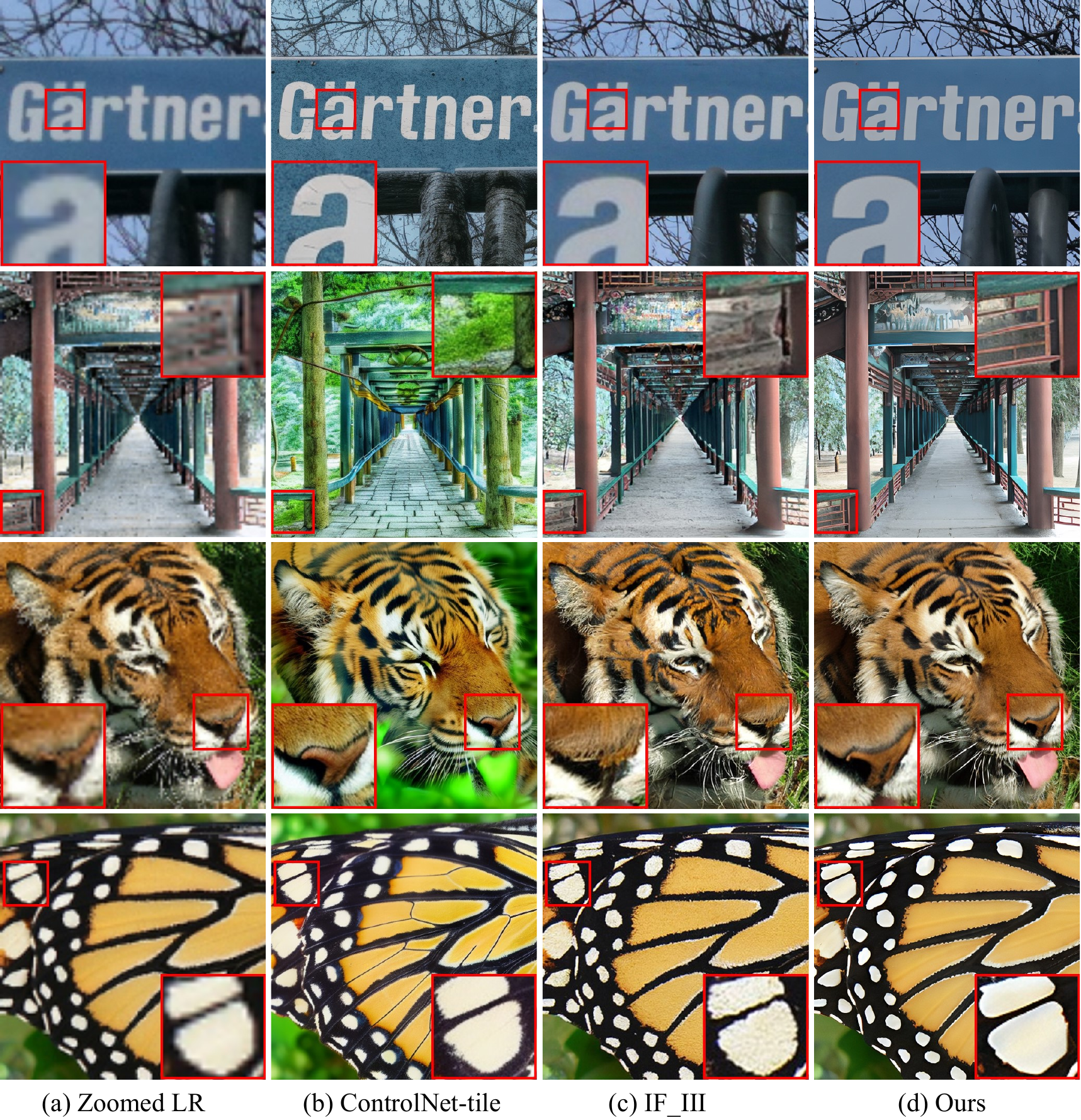}}
\caption{Qualitative comparisons on real-world images ($128 \rightarrow 512$).
Our StableSR outperforms ControlNet-tile \citep{zhang2023adding} with higher fidelity and has more realistic and sharper details compared with IF\_III upscaler \citep{deepfloyd2023}. (\textbf{Zoom in for details})
}
\label{fig_compare}
\vspace{-0.7cm}
\end{center}
\end{figure}

\begin{figure}[!ht]
\begin{center}
\centerline{\includegraphics[width=1.0\columnwidth]{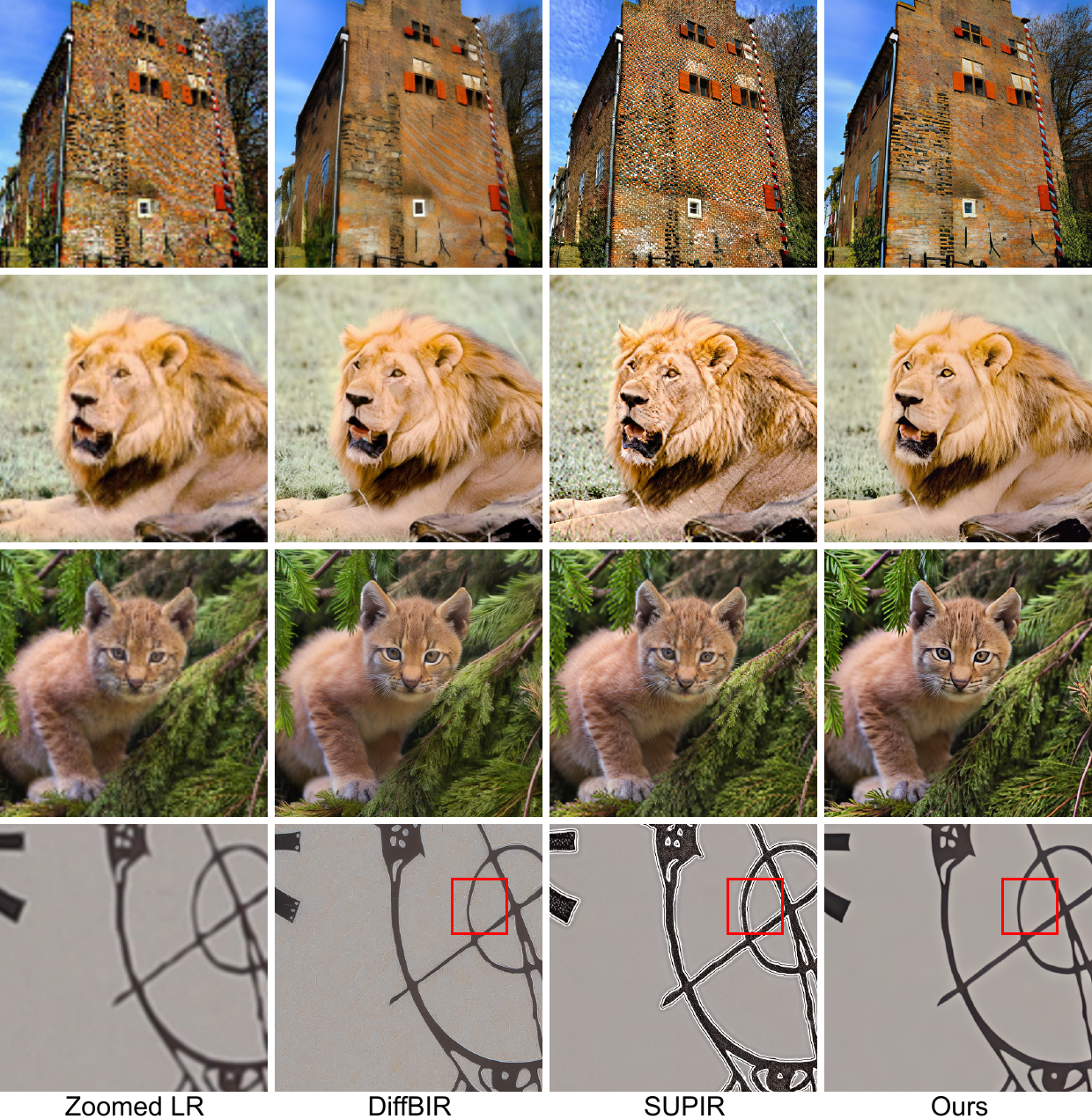}}
\vspace{-0.1cm}
\caption{Qualitative comparisons on real-world images ($128 \rightarrow 512$) with DiffBIR~\citep{lin2023diffbir} and SUPIR~\citep{yu2024scaling}. 
(\textbf{Zoom in for details})
}
\label{fig:follow}
\vspace{-0.7cm}
\end{center}
\end{figure}

\noindent\textbf{Comparison with Concurrent Diffusion Applications.}
We notice that recent concurrent works \citep{zhang2023adding,deepfloyd2023} can also be adopted for image SR.
While IF\_III upscaler \citep{deepfloyd2023} is a super-resolution model training from scratch, ControlNet-tile \citep{zhang2023adding} also adopts a diffusion prior.
The key technical differences regarding to the use of diffusion prior between our StableSR and ControlNet-tile lie in the different adaptor designs, i.e., ControlNet-tile adopts a trainable copy of the encoding layers in Stable Diffusion~\citep{rombach2021highresolution}, whilst StableSR does not rely on any layer copies of the fixed diffusion prior, thus can be more flexible.
Specifically, we introduce a time-aware encoder to modulate the feature maps of the fixed diffusion prior. This time-aware encoder is more lightweight than the copied layers in ControlNet-tile, i.e., 105M vs. 364M. As a result, StableSR is also faster than ControlNet-tile in terms of inference speed, i.e., 10.37s vs. 14.47s for 50 sampling steps.
Here, we further conduct comparisons with these methods on real-world images.
For fair comparisons, we use DDIM sampling with $\eta=1.0$ and timestep $200$ for all the methods, and the seed is fixed to $42$.
We further set $w = 0.0$ in StableSR to avoid additional improvement due to CFW.
For ControlNet-tile \citep{zhang2023adding}, we generate additional prompts using stable-diffusion-webui\footnote{https://github.com/AUTOMATIC1111/stable-diffusion-webui} for better performance.
For IF\_III upscaler \citep{deepfloyd2023}, we follow official examples to set noise level to $100$ w/o prompts.
As shown in Fig.~\ref{fig_compare}, ControlNet-tile shows poor fidelity due to the lack of specific designs for SR.
Compared with IF\_III upscaler, the proposed StableSR is capable of generating more faithful details with sharper edges, \eg, the text in the first row, the tiger's nose in the third row and the wing of the butterfly in the last row of Fig.~\ref{fig:follow}.
Note that IF\_III upscaler is trained from scratch, which requires significant computational resources.
The visual comparisons suggest the superiority of StableSR.

\noindent\textbf{Comparison with Follow-up Approaches.}
During the submission of our work, we notice that several follow-up methods~\citep{lin2023diffbir,yu2024scaling} are further proposed for image super-resolution by exploiting the diffusion prior with a ControlNet-like~\citep{zhang2023adding} framework.
We therefore conduct a further comparison with these works here.
The key technical differences regarding the use of diffusion prior between our StableSR and DiffBIR lie in the different adaptor designs, i.e., DiffBIR follows ControlNet~\citep{zhang2023adding} to adopt a trainable copy of the encoding layers in Stable Diffusion~\citep{rombach2021highresolution}, while StableSR does not rely on any layer copies of the fixed diffusion prior, thus can be more flexible.
Specifically, the generation module part of DiffBIR is the same as ControlNet, leading to more trainable parameters (364M vs. 105M) and longer inference time (14.47s vs. 10.37s).
Besides, DiffBIR requires an additional pre-clean model during both training and inference, as inspired by our earlier work DifFace~\citep{yue2022difface}, whilst our StableSR does not require such a pre-clean model during training. In the testing phase, this pre-clean model is also optional and can be removed\footnote{We do not use it by default, unless clarified.}. Details of the pre-clean model for StableSR can be found in the supplementary material.
Similar to DiffBIR, another recent work SUPIR~\citep{yu2024scaling} proposes to adopt SDXL~\citep{podell2023sdxl}, a much larger diffusion model (2.6B vs. 865M) as diffusion prior and develops a trimmed ControlNet to reduce the model size.
While both following ControlNet~\citep{zhang2023adding}, SUPIR has much more trainable parameters, i.e., 1.3B than DiffBIR, leading to almost 2x inference time than StableSR.
We further conduct comparisons on real-world test data.
As shown in Table~\ref{tab:followup} and Fig.~\ref{fig:follow}, StableSR is comparable with DiffBIR. We further notice that DiffBIR tends to generate patterns overly
as shown in the last row of Fig.~\ref{fig:follow} while StableSR does not suffer from such a problem.
As for SUPIR, we observe that it does not perform well on images with small resolutions, i.e., lower than 512 after upsampling.
We conjecture this is because small cropped images lack semantic content and the prior adopted by SUPIR is trained on a $1024 \times 1024$ resolution.
However, we do observe that SUPIR outperforms our method on large resolutions beyond $1024$, which should be mostly due to the huge model size and the large training set with detailed prompts.
Improving StableSR with larger diffusion prior and training datasets with prompts can be regarded as a future direction.

\begin{table}[!t]
 \newcommand{\tabincell}[2]{\begin{tabular}{@{}#1@{}}#2\end{tabular}}
 \begin{center}
  \caption{Quantitative comparison with follow-up works, i.e., DiffBIR~\citep{lin2023diffbir} and SUPIR~\citep{yu2024scaling} on RealSR \citep{cai2019toward} and DRealSR \citep{wei2020component} benchmarks. SUPIR does not perform well due to the resolution gap between test data ($512 \times 512$) and SDXL prior ($1024 \times 1024$).}
  \label{tab:followup}
\resizebox{0.5\textwidth}{!}{
  \begin{tabular}{c|c|ccc}
        \hline Datasets & Metrics & \makecell[c]{DiffBIR} &  \makecell[c]{SUPIR} & \makecell[c]{\textbf{StableSR}} \\
        \hline \multirow{5}{*}{\makecell[c]{RealSR}}
        & PSNR $\uparrow$ & \textbf{25.02} & 23.70 & 24.65 \\
         & SSIM $\uparrow$ & 0.6711 & 0.6647 & \textbf{0.7080} \\
         & LPIPS $\downarrow$ & 0.3568 & 0.3559 & \textbf{0.3002} \\
         & CLIP-IQA $\uparrow$ & 0.6568 & \textbf{0.6619} & 0.6234 \\
        & MUSIQ $\uparrow$ & 64.07 & 61.97 & \textbf{65.88} \\
        \hline \multirow{5}{*}{\makecell[c]{DRealSR}}
        & PSNR $\uparrow$ & 27.20 & 24.86 & \textbf{28.03} \\
         & SSIM $\uparrow$ & 0.6721 & 0.6441 & \textbf{0.7536} \\
         & LPIPS $\downarrow$ & 0.4274 & 0.4229 & \textbf{0.3284} \\
        & CLIP-IQA $\uparrow$ & 0.6293 & \textbf{0.6891} & 0.6357 \\
        & MUSIQ $\uparrow$ & \textbf{59.87} & 59.70 & 58.51 \\
        \hline
  \end{tabular}
}
\vspace{-0.5cm}
 \end{center}
\end{table}

\vspace{-0.2cm}
\subsection{Ablation Study}
\label{sec:abla}

\begin{figure*}[htbp]
\begin{center}
\centerline{\includegraphics[width=2.0\columnwidth]{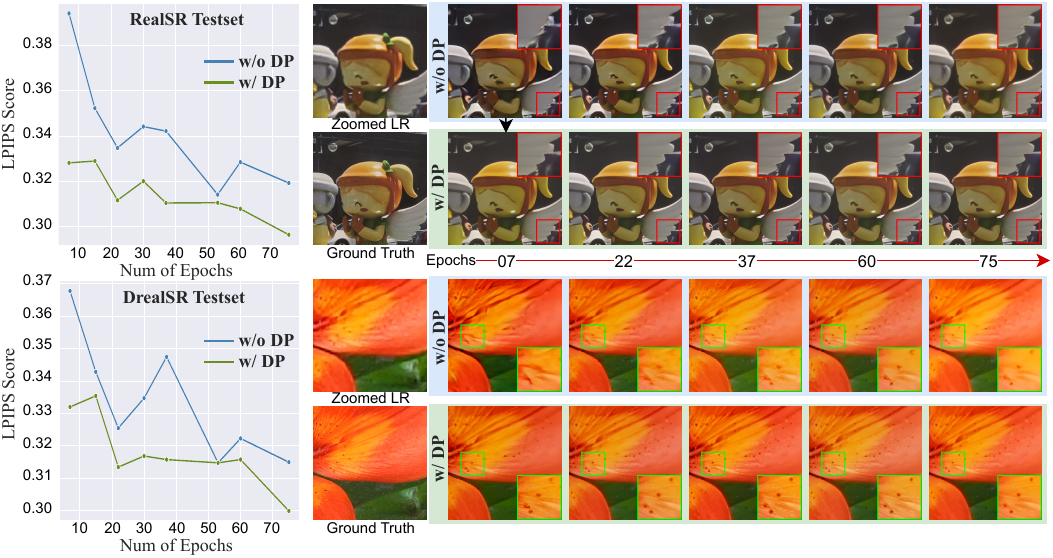}}
\vspace{-0.1cm}
\caption{Training comparisons between w/ and w/o diffusion prior (DP). Adopting DP significantly speeds up the training process with better LPIPS scores at early epochs. The visualization results on validation sets at different epochs also indicate the superiority of using DP. (\textbf{Zoom in for details})}
\label{fig:noprior}
\end{center}
\vspace{-0.7cm}
\end{figure*}

\noindent\textbf{Effectiveness of Diffusion Prior.}
We first verify the effectiveness of adopting diffusion prior for super-resolution.
We train a baseline from scratch without loading a pretrained diffusion model as diffusion prior.
The architecture is kept the same as our StableSR for fair comparison.
As shown in Fig.~\ref{fig:noprior}, benefiting from the diffusion prior, StableSR achieves better LPIPS scores on both of the validation datasets during training.
The visual comparisons at different epochs also indicate the significance of adopting diffusion prior.
Moreover, we observe that training from scratch requires 2.06 times more GPU memory in average compared to StableSR on NVIDIA Tesla 32G-V100 GPUs.

\begin{figure*}[htbp]
\begin{center}
\centerline{\includegraphics[width=2.0\columnwidth]{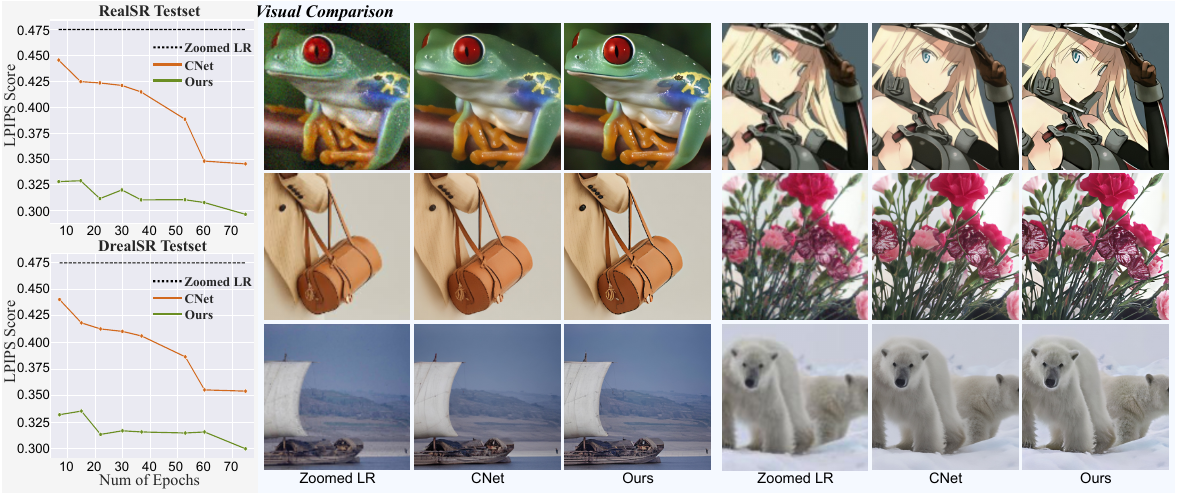}}
\vspace{-0.1cm}
\caption{Training process and qualitative comparisons with ControlNet (CNet) baseline~\citep{zhang2023adding}. As shown in the validation curve during training, StableSR converges faster with better LPIPS scores on both validation sets. The visual comparisons after training 117 epochs also indicate the effectiveness of our StableSR. (\textbf{Zoom in for details})}
\label{fig:cnet}
\end{center}
\vspace{-0.8cm}
\end{figure*}

\begin{figure}[htbp]
\begin{center}
\centerline{\includegraphics[width=1.0\columnwidth]{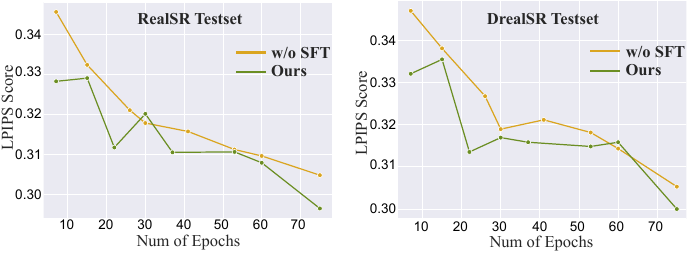}}
\caption{Training comparisons between our StableSR and the baseline w/o SFT layers. SFT layers slightly improve the training performance in terms of lower LPIPS scores of validation sets.}
\label{fig:nosft}
\end{center}
\vspace{-0.7cm}
\end{figure}

\noindent\textbf{Effectiveness of Network Design.}
In StableSR, a time-aware encoder and SFT layers are adopted to harness the diffusion prior.
While concurrent works ControlNet~\citep{zhang2023adding} and T2I-Adaptor~\citep{mou2023t2i} propose to exploit diffusion prior to image generation, their effectiveness for image super-resolution is underexplored.
Here, we further compare our design with theirs.
Specifically, we first retrain a ControlNet for image super-resolution using the same diffusion prior and training pipelines as ours.
Recall that we have shown the superiority of StableSR compared with ControlNet-tile in Fig.~\ref{fig_compare}.
With retraining, the performance of ControlNet for super-resolution can be improved, but still inferior to ours as shown in Fig.~\ref{fig:cnet}.
To compare with T2I-Adapter, while we have already verified the effectiveness of time-aware guidance, we further add a baseline w/o SFT layers by first mapping the features to the same shape as the prior features and then adding them together.
Note that such strategy can be regarded as a special case of SFT layers with $\bm{\alpha}^n=0, \bm{\beta}^n=0$ in Eq.\eqref{eq:sft}.
As shown in Fig.~\ref{fig:nosft}, SFT layers slightly improve the training performance on the validation sets in terms of LPIPS scores during training.

\begin{figure}[!h]
\begin{center}
\centerline{\includegraphics[width=1.0\columnwidth]{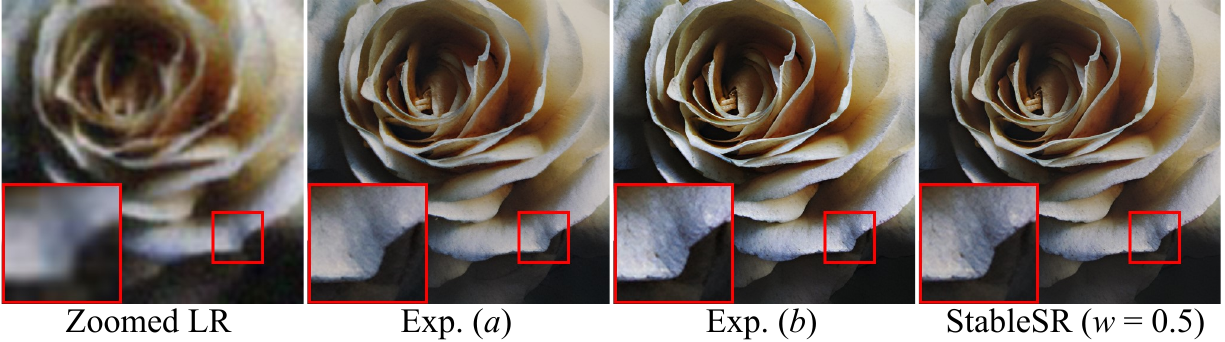}}
\caption{Visual comparisons of time-aware guidance and color correction. Exp. (a) does not apply time-aware guidance, leading to blurry textures. Exp. (b) applies time-aware guidance and can generate sharper details, but obvious color shifts can be observed. With both strategies, StableSR generates sharp textures and avoids color shifts.
}
\label{abla_t_color}
\vspace{-0.8cm}
\end{center}
\end{figure}

\begin{figure}[!t]
\begin{center}
\centerline{\includegraphics[width=1.0\columnwidth]{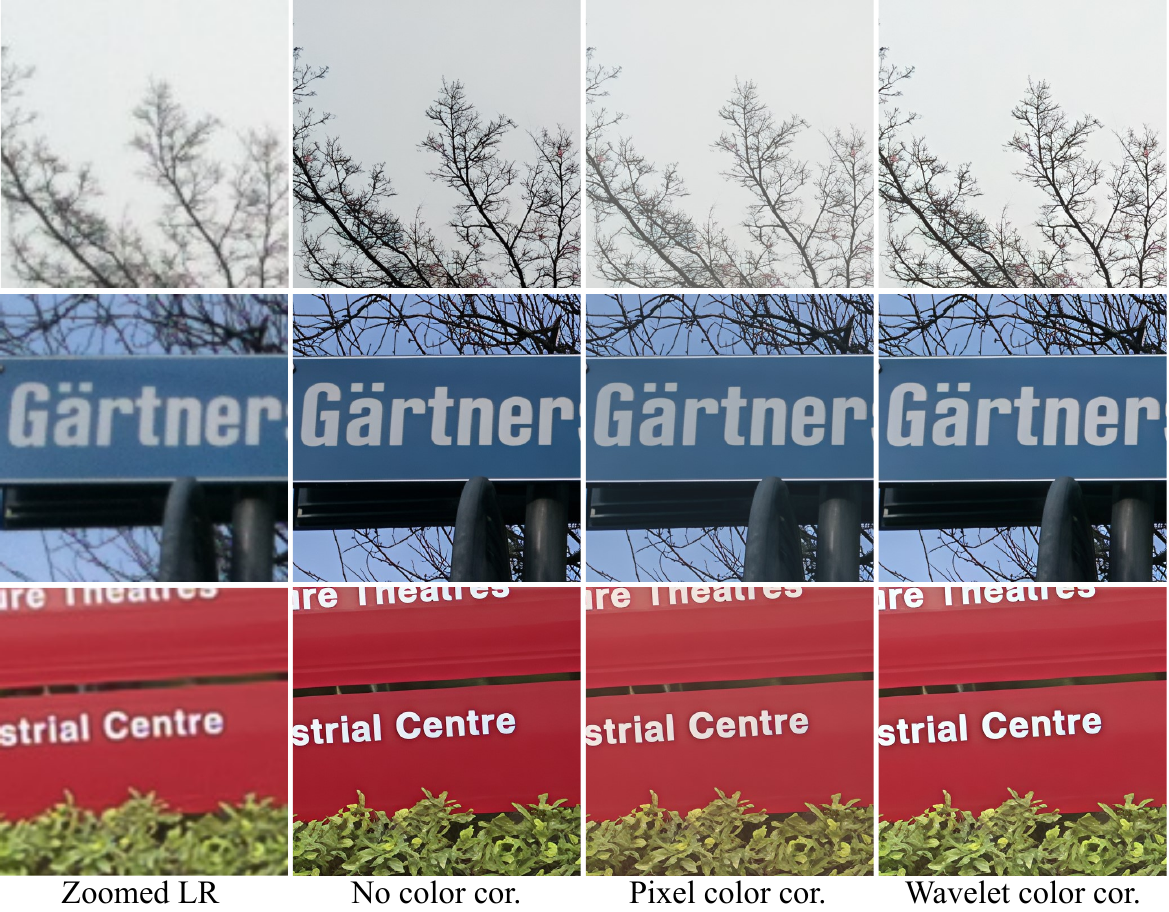}}
\caption{Visual comparisons of different color correction strategies. With no color correction, obvious color shifts can be observed in Exp. (b). Our color correction via channel matching in Eq.~\eqref{eq:color_chn_matching} can alleviate the color shift problem, while the wavelet color correction of Eq.~\eqref{eq:color_wavlet} can further improve the visual quality in these cases.
}
\label{abla_t_wavelet}
\vspace{-0.5cm}
\end{center}
\end{figure}

\noindent\textbf{Importance of Time-aware Guidance and Color Correction.}
We then investigate the significance of time-aware guidance and color correction.
Recall that in Fig.~\ref{fea_cos}, we already show that the time-aware guidance allows the encoder to adaptively adjust the condition strength.
Here, we further verify its effectiveness on real-world benchmarks \citep{cai2019toward,wei2020component}.
As shown in Table~\ref{tab:abla}, removing time-aware guidance (\ie,~removing the time-embedding layer) or color correction both lead to worse SSIM and LPIPS.
Moreover, the comparisons in Fig.~\ref{abla_t_color} also indicate inferior performance without the above two components, suggesting the effectiveness of time-aware guidance and color correction.
In addition to directly adopting color correction in the pixel domain, our proposed wavelet color correction can further boost the visual quality, as shown in Fig. \ref{abla_t_wavelet}, which may further facilitate the practical use.
Note that technically, the wavelet transform may introduce halo effects~\citep{thorndike1920constant}, though we do not observe this phenomenon during our experiments.

\begin{table}[!t]
 \footnotesize
 \newcommand{\tabincell}[2]{\begin{tabular}{@{}#1@{}}#2\end{tabular}}
 \begin{center}
  \caption{Ablation studies of time-aware guidance and color correction on RealSR \citep{cai2019toward} and DRealSR \citep{wei2020component} benchmarks.}
  \label{tab:abla}
\resizebox{0.5\textwidth}{!}{
  \begin{tabular}{c|c|c|c|ccc}
  \hline  \multirow{3}{*}{Exp.} & \multicolumn{3}{c|}{Strategies} & \multicolumn{3}{c}{RealSR / DRealSR} \\
  \cline{2-7} & \makecell[c]{Time \\ aware} & \makecell[c]{Pixel \\Color cor.} & \makecell[c]{Wavelet \\ Color cor.}  & PSNR $\uparrow$ & SSIM $\uparrow$ & LPIPS $\downarrow$ \\
  \hline (a) &  & \cmark & & \textbf{24.65} / 27.68 & 0.7040 / 0.7280 & 0.3157 / 0.3456 \\
   (b) & \cmark & & & 22.24 / 23.86 & 0.6840 / 0.7179 & 0.3180 / 0.3544 \\
   (c) & \cmark & & \cmark & 23.38 / 26.80 & 0.6870 / 0.7235 & 0.3157 / 0.3475 \\
   Default & \cmark & \cmark & & \textbf{24.65} / \textbf{28.03} & \textbf{0.7080} / \textbf{0.7536} & \textbf{0.3002} / \textbf{0.3284} \\
            \hline
  \end{tabular}
}
 \end{center}
\vspace{-0.5cm}
\end{table}

\begin{figure}[!ht]
\begin{center}
\centerline{\includegraphics[width=1.0\columnwidth]{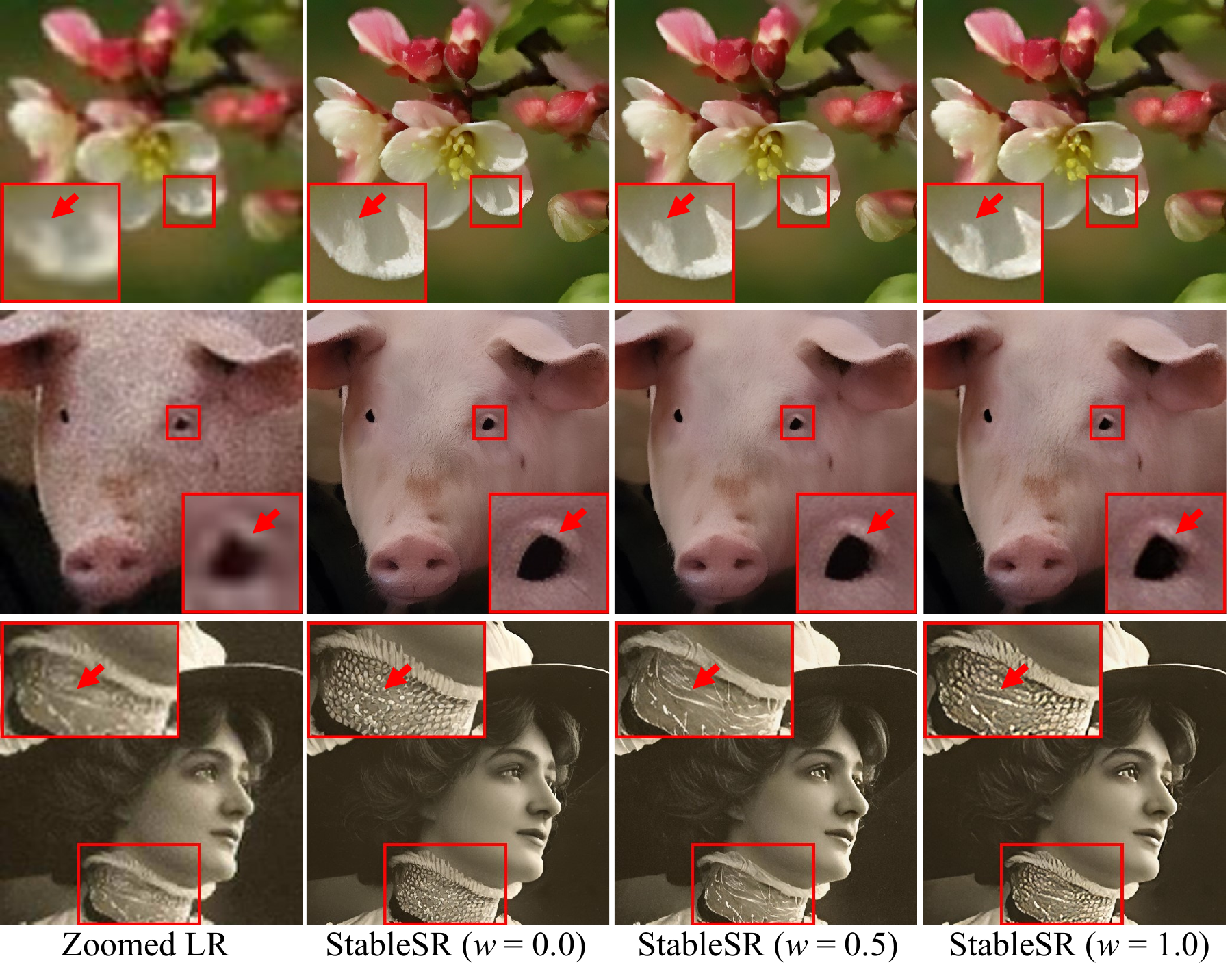}}
\vspace{-0.1cm}
\caption{Visual comparisons with different coefficients $w$ for CFW module. A small $w$ tends to generate a realistic result while a larger $w$ improves the fidelity.
}
\label{abla_w}
\vspace{-0.8cm}
\end{center}
\end{figure}

\noindent\textbf{Flexibility of Fidelity-realism Trade-off.}
Our CFW module inspired by CodeFormer~\citep{zhou2022codeformer} allows a flexible realism-fidelity trade-off.
In particular, given a controllable coefficient $w$ with a range of $[0,1]$, CFW with a small $w$ tends to generate a realistic result, especially for large degradations, while CFW with a larger $w$ improves the fidelity.
As shown in Table \ref{metric_weight}, compared with StableSR ($w=0.0$), StableSR with larger values of $w$ (e.g., 0.75) achieves higher PSNR and SSIM on all three paired benchmarks, indicating better fidelity.
In contrast, StableSR ($w=0.0$) achieves better perceptual quality with higher CLIP-IQA scores and MUSIQ scores.
Similar phenomena can also be observed in Fig.~\ref{abla_w}.
We further observe that a proper $w$ can lead to improvement in both fidelity and perceptual quality.
Specifically, StableSR ($w=0.5$) shows comparable PSNR and SSIM with StableSR ($w=1.0$) but achieves better perceptual metric scores in Table \ref{metric_weight}.
Hence, we set the coefficient $w$ to 0.5 by default for trading between quality and fidelity.
We observe that CFW necessitates extra GPU memory.
Consequently, we designate it as an optional feature for varying applications.

\begin{table}[!t]
 \newcommand{\tabincell}[2]{\begin{tabular}{@{}#1@{}}#2\end{tabular}}
 \begin{center}
  \caption{Ablation studies of the controllable coefficient $w$ on both synthetic (DIV2K Valid \citep{agustsson2017ntire}) and real-world (RealSR \citep{cai2019toward}, DRealSR \citep{wei2020component}, and DPED-iPhone \citep{ignatov2017dslr}) benchmarks.}
  \label{metric_weight}
\resizebox{0.5\textwidth}{!}{
  \begin{tabular}{c|c|cccc}
                \hline Datasets & Metrics & \makecell[c]{StableSR \\ ($w=0.0$)} & \makecell[c]{\textbf{StableSR} \\ ($w=0.5$)} & \makecell[c]{StableSR \\ ($w=0.75$)} & \makecell[c]{StableSR \\ ($w=1.0$)} \\
                \hline \multirow{6}{*}{\makecell[c]{DIV2K Valid}} & PSNR $\uparrow$ & 22.68 & 23.26 & \textbf{24.17} & 23.14 \\
                 & SSIM $\uparrow$ & 0.5546 & 0.5726 & \textbf{0.6209} & 0.5681 \\
                 & LPIPS $\downarrow$ & 0.3393 & 0.3114 & \textbf{0.3003} & 0.3077 \\
                & FID $\downarrow$ & 25.83 & 24.44 & \textbf{24.05} & 26.14 \\
                & CLIP-IQA $\uparrow$ & 0.6529 & \textbf{0.6771} & 0.5519 & 0.6197\\
                & MUSIQ $\uparrow$ & 65.72 & \textbf{65.92} & 59.46 & 64.31 \\
                \hline
                \hline \multirow{5}{*}{\makecell[c]{RealSR}}
                & PSNR $\uparrow$ & 24.07 & 24.65 & \textbf{25.37} & 24.70 \\
                 & SSIM $\uparrow$ & 0.6829 & 0.7080 & \textbf{0.7435} & 0.7157 \\
                 & LPIPS $\downarrow$ & 0.3190 & 0.3002 & \textbf{0.2672} & 0.2892 \\
                 & CLIP-IQA $\uparrow$ & 0.6127 & \textbf{0.6234} & 0.5341 & 0.5847\\
                & MUSIQ $\uparrow$ & 65.81 & \textbf{65.88} & 62.36 & 64.05\\
                \hline \multirow{5}{*}{\makecell[c]{DRealSR}}
                & PSNR $\uparrow$ & 27.43 & 28.03 & \textbf{29.00} & 27.97 \\
                 & SSIM $\uparrow$ & 0.7341 & 0.7536 & \textbf{0.7985} & 0.7540 \\
                 & LPIPS $\downarrow$ & 0.3595 & 0.3284 & \textbf{0.2721} & 0.3080 \\
                & CLIP-IQA $\uparrow$ & 0.6340 & \textbf{0.6357} & 0.5070 & 0.5893\\
                & MUSIQ $\uparrow$ & \textbf{58.98} & 58.51 & 53.12 & 56.77 \\
                \hline \multirow{2}{*}{\makecell[c]{DPED-iPhone}} & CLIP-IQA $\uparrow$ & \textbf{0.5015} & 0.4799 & 0.3405 & 0.4250\\
                & MUSIQ $\uparrow$ & \textbf{51.90} & 50.48 & 41.81 & 47.96\\
                \hline
  \end{tabular}
}
\vspace{-0.5cm}
 \end{center}
\end{table}

\begin{table*}[ht]
 \newcommand{\tabincell}[2]{\begin{tabular}{@{}#1@{}}#2\end{tabular}}
 \begin{center}
  \caption{Complexity comparison of model complexity. All methods are evaluated on $128 \times 128$ input images for 4x SR using an NVIDIA Tesla 32G-V100 GPU. The runtime is averaged by ten runs with a batch size of 1.}
  \label{metric_runtime}
\resizebox{1.0\textwidth}{!}{
  \begin{tabular}{c|c|c|c|c|c|c|c}
                \hline & \makecell[c]{Real-ESRGAN+} & \makecell[c]{FeMaSR} & \makecell[c]{SwinIR-GAN} & \makecell[c]{LDM} & \makecell[c]{IF\_III} & \makecell[c]{\textbf{StableSR}} & \makecell[c]{\textbf{StableSR-Turbo}}\\
                \hline
                \makecell[c]{Model type} & GAN & GAN & GAN & Diffusion & Diffusion & Diffusion & Diffusion\\
                \hline
                \makecell[c]{Number of \\ Inference step} & 1 & 1 & 1 & 200 & 200 & 200 & 4 \\
                \hline
                Runtime & 0.08s & 0.12s & 0.31s & 5.25s & 17.78s & 15.16s & 0.83s\\
                \hline
                \makecell[c]{Trainable \\ Params} & 16.70M & 28.29M & 28.01M & 113.62M & 473.40M & 149.91M & 149.91M \\
                \hline
  \end{tabular}
}
\vspace{-0.5cm}
 \end{center}
\end{table*}

\vspace{-0.2cm}
\subsection{Complexity Comparison}
StableSR is a diffusion-based approach and requires multi-step sampling for image generation.
As shown in Table~\ref{metric_runtime}, when the number of sampling steps is set to 200, StableSR needs 15.16 seconds to generate a $512 \times 512$ image on one NVIDIA Tesla 32G-V100 GPU.
This is comparable to IF\_III upscaler but slower than GAN-based SR methods such as Real-ESRGAN+ and SwinIR-GAN, which require only a single forward pass.
Fast sampling strategy \citep{song2020denoising,lu2022dpm,karras2022elucidating} and model distillation \citep{salimans2021progressive,song2023consistency,luo2023latent} are two promising solutions to improve efficiency. Another viable remedy is to shorten the chain of diffusion process \citep{yue2023resshift}.
As for trainable parameters, StableSR has $149.91$M trainable parameters, which is only 11.50\% of the full model and less than IF\_III, \ie, 473.40M.
The trainable parameters can be further decreased with more careful design, \eg, adopting lightweight architectures \citep{chollet2017xception,howard2019searching} or network pruning \citep{fang2023structural}.
Such exploration is beyond the scope of this paper.

\vspace{-0.1cm}
\section{Inference Strategies}
The proposed StableSR already demonstrates superior performance quantitatively and qualitatively on both synthetic and real-world benchmarks, as shown in Sec.~\ref{sec_exp}. Here, we discuss several effective strategies during the sampling process that can further boost the inference performance without additional finetuning.

\begin{figure}[!t]
\begin{center}
\centerline{\includegraphics[width=1.0\columnwidth]{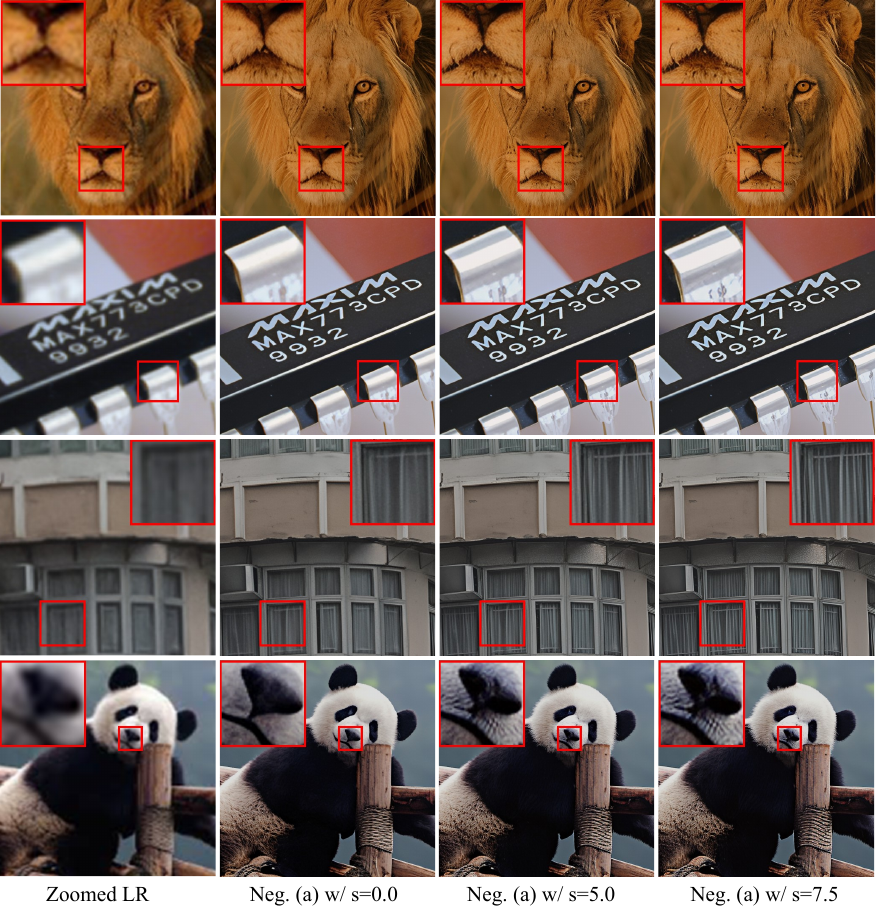}}
\vspace{-0.1cm}
\caption{Qualitative comparisons on classifier-free guidance with negative prompts.
Higher guidance scale $s$ leads to sharper edges. (\textbf{Zoom in for details})
}
\label{fig_classscale}
\vspace{-0.7cm}
\end{center}
\end{figure}

\vspace{-0.1cm}
\subsection{Classifier-free Guidance with Negative Prompts}
The default StableSR is trained with null prompts. Interestingly, we observe that StableSR can react to prompts, especially negative prompts.
We examine the use of classifier-free guidance \citep{ho2021classifier} with negative prompts to further improve the visual quality during sampling.
Given two StableSR models conditioned on null prompts $\epsilon_{\bm{\theta}}(\bm{Z}^{(t)}, \bm{F}, [], t)$ and negative prompts $\epsilon_{\bm{\theta}}(\bm{Z}^{(t)}, \bm{F}, \bm{c}, t)$, respectively, the new sampling process can be performed using a linear combination of the estimations with a guidance scale $s$:
\begin{equation}
\label{eq:guide}
\resizebox{0.43\textwidth}{!}{
    $\tilde{\epsilon}_{\bm{\theta}} = \epsilon_{\bm{\theta}}(\bm{Z}^{(t)}, \bm{F}, \bm{c}, t) + s\left(\epsilon_{\bm{\theta}}(\bm{Z}^{(t)}, \bm{F}, [], t) - \epsilon_{\bm{\theta}}(\bm{Z}^{(t)}, \bm{F}, \bm{c}, t)\right)$},
\end{equation}
where $\bm{c}$ is the negative prompt for guidance.
According to Eq. \eqref{eq:guide}, it is worth noting that $s=0$ is equivalent to directly using negative prompts without guidance, and $s=1$ is equivalent to our default settings with the null prompt.

We compare the performance of StableSR with various positive prompts, \ie, (1) ``\texttt{(masterpiece:2), (best quality:2), (realistic:2), (very clear:2)}'', and (2) ``\texttt{Good photo.}'', and negative prompts, \ie, (a) ``\texttt{3d, cartoon, anime, sketches, (worst quality:2), (low quality:2)}'', and (b) ``\texttt{Bad photo.}''.
As shown in Table~\ref{tab:prompt}, different prompts lead to diverse metric scores.
Specifically, the classifier-free guidance with negative prompts shows a significant influence on the metrics, \ie, higher guidance scales lead to higher CLIP-IQA and MUSIQ scores, indicating sharper results.
Similar phenomena can also be observed in Fig.~\ref{fig_classscale}.
However, an overly strong guidance, \eg, $s=7.5$ can result in oversharpening.

\begin{figure*}[htbp]
\begin{center}
\centerline{\includegraphics[width=2.05\columnwidth]{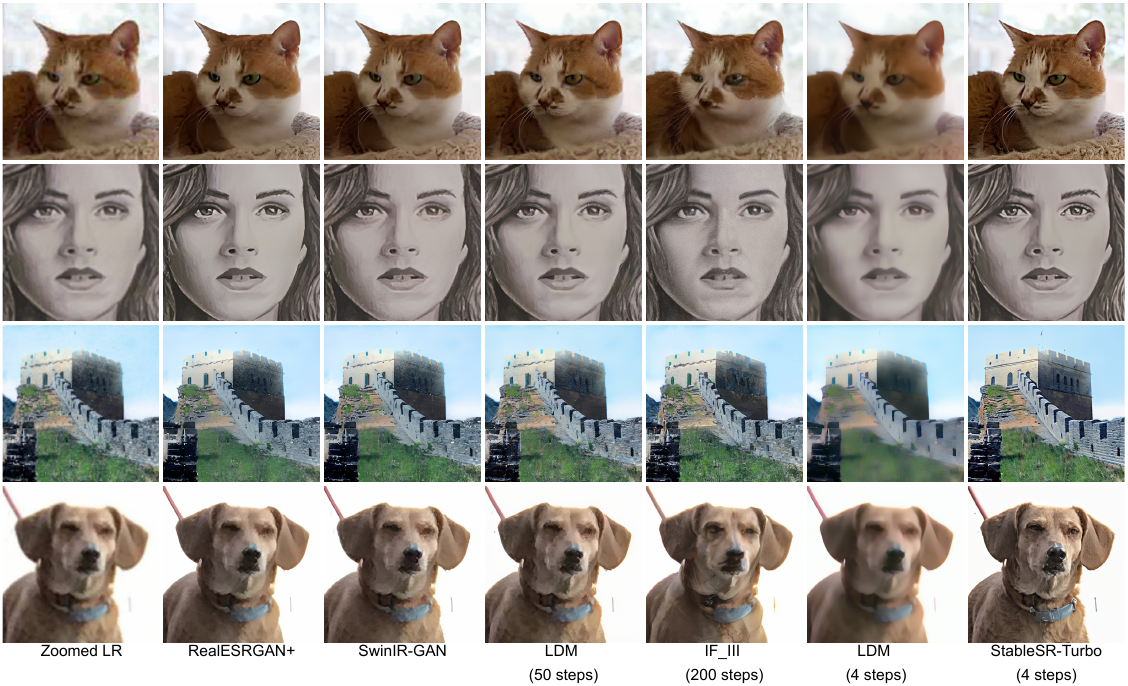}}
\vspace{-0.20cm}
\caption{Qualitative comparisons on real-world images ($128 \rightarrow 512$). Our StableSR-Turbo w/o further finetuning is capable of generating high-quality images in only 4 steps, while still significantly outperforming existing approaches.}
\label{fig_turbo}
\end{center}
\vspace{-0.8cm}
\end{figure*}

\begin{table*}[htbp]
 \newcommand{\tabincell}[2]{\begin{tabular}{@{}#1@{}}#2\end{tabular}}
 \begin{center}
 \vspace{-0.1cm}
  \caption{Comparison of different prompts and guidance strengths. Note that $s=0$ is equivalent to using negative prompts w/o guidance. Positive prompts are (1) ``\texttt{(masterpiece:2), (best quality:2), (realistic:2), (very clear:2)}'', and (2) ``\texttt{Good photo.}''. Negative prompts are (a) ``\texttt{3d, cartoon, anime, sketches, (worst quality:2), (low quality:2)}'', and (b) ``\texttt{Bad photo.}''. The first row is the default settings for StableSR.}
  \label{tab:prompt}
  \begin{tabular}{c|c|c|ccccc}
  \hline  \multicolumn{3}{c|}{Strategies} & \multicolumn{4}{c}{RealSR / DRealSR} \\
  \hline \makecell[c]{Pos. \\ Prompts} & \makecell[c]{Neg. \\ Prompts} & \makecell[c]{Guidance \\ Scale}  & PSNR $\uparrow$ & SSIM $\uparrow$ & LPIPS $\downarrow$ & CLIP-IQA $\uparrow$ & MUSIQ $\uparrow$ \\
  \hline [] & - & - & 24.65 / 28.03 & 0.7080 / 0.7536 & 0.3002 / 0.3284 & 0.6234 / 0.6357 & 65.88 / 58.51 \\
  \hline (1) & - & - & 24.68 / 28.03 & 0.7025 / 0.7461 & 0.3151 / 0.3378 & 0.6251 / 0.6370 & 65.34 / 58.07 \\
  \hline (2) & - & - & 24.71 / 28.07 & 0.7049 / 0.7500 & 0.3118 / 0.3333 & 0.6219 / 0.6291 & 65.22 / 57.75 \\
  \hline \multirow{4}{*}{\makecell[c]{[]}} & \multirow{4}{*}{\makecell[c]{(a)}} & $s=0.0$ & 24.80 / 28.18 & 0.7097 / 0.7562 & 0.3105 / 0.3316 & 0.6176 / 0.6224 &64.86 / 57.31 \\
  \cline{3-8} & & $s=2.5$ &24.41 / 27.76 &0.6972 / 0.7383 & 0.3168 / 0.3417 & 0.6306 / 0.6422 & 66.02 / 59.21 \\
  \cline{3-8} & & $s=5.0$ & 23.96 / 27.21 &0.6829 / 0.7188 &0.3267 / 0.3583 & 0.6356 / 0.6558 &66.84 / 61.07 \\
  \cline{3-8} & & $s=7.5$ & 23.53 / 26.68 &0.6673 / 0.7003 & 0.3399 / 0.3774 &0.6323 / 0.6621 &67.26 / 62.41 \\
  \hline \multirow{4}{*}{\makecell[c]{[]}} & \multirow{4}{*}{\makecell[c]{(b)}} & $s=0.0$ &24.77 / 28.13 &0.7067 / 0.7520 & 0.3100 / 0.3317 &0.6184 / 0.6239 &64.81 / 57.27 \\
  \cline{3-8} & & $s=2.5$ &24.46 / 27.90 &0.7017 / 0.7467 & 0.3170 / 0.3371 &0.6303 / 0.6409 &66.29 / 58.97 \\
  \cline{3-8} & & $s=5.0$ &24.13 / 27.61 &0.6958 / 0.7391 & 0.3240 / 0.3467 &0.6377 / 0.6490 &67.43 / 60.69 \\
  \cline{3-8} & & $s=7.5$ &23.78 / 27.30 &0.6894 / 0.7310 & 0.3320 / 0.3578 & 0.6421 / 0.6583 & 68.13 / 62.12 \\
  \hline
  \end{tabular}
 \end{center}
 \vspace{-0.45cm}
\end{table*}

\begin{figure*}[!ht]
\begin{center}
\centerline{\includegraphics[width=2.05\columnwidth]{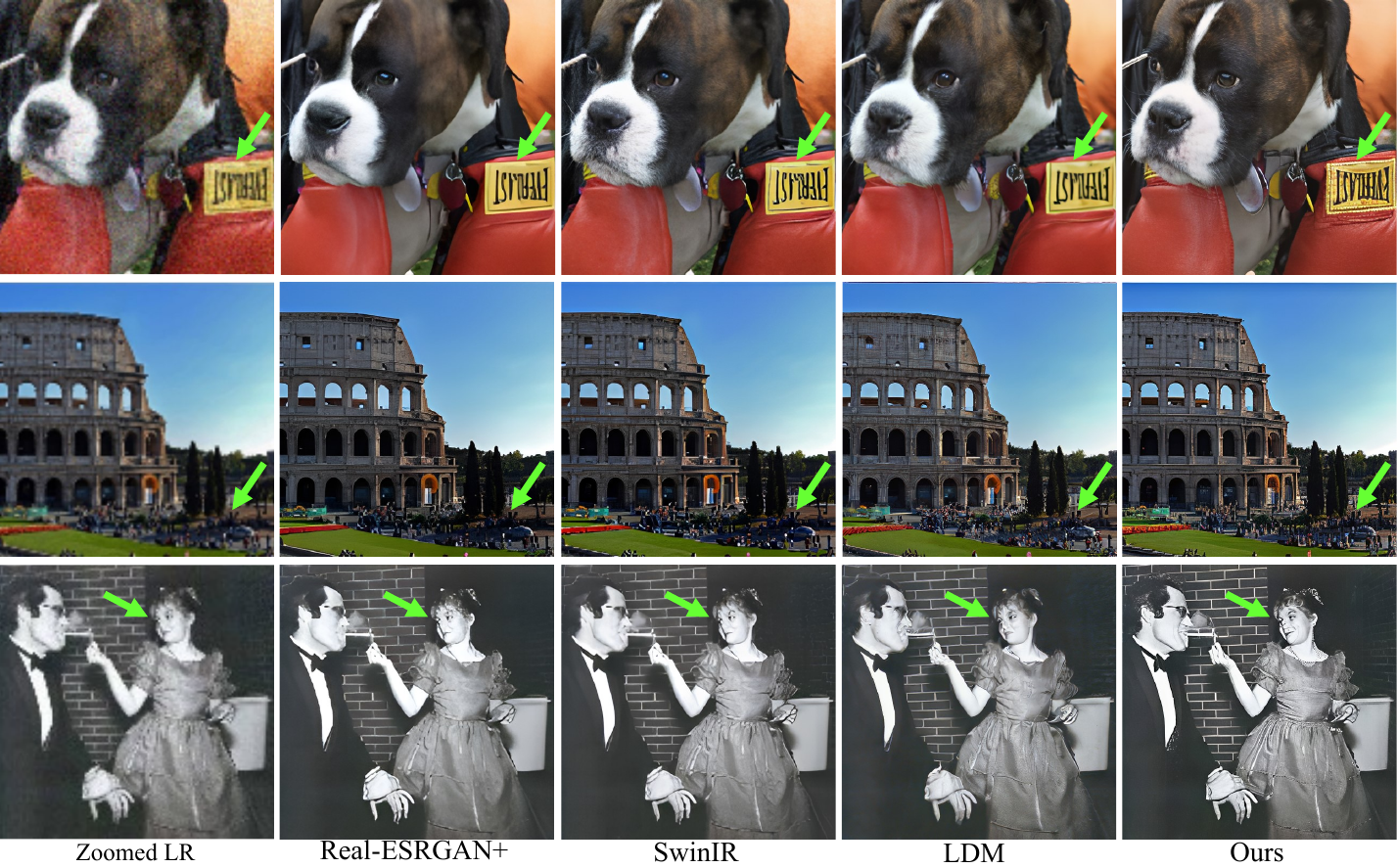}}
\vspace{-0.20cm}
\caption{StableSR shares the same limitations as the diffusion prior, i.e., Stable Diffusion~
\citep{rombach2021highresolution}, thus may fail to handle texts, very small patterns and small faces. While these cases are very challenging for existing generic SR methods, we believe a more powerful diffusion prior and larger data training could help.}
\label{fig:failcase}
\end{center}
\vspace{-0.6cm}
\end{figure*}

\subsection{StableSR with SD-Turbo}
The default sampler of StableSR is DDPM~\citep{ho2020denoising} with 200 sampling steps.
Though effective, the sampling process can be time-consuming compared with non-diffusion approaches as shown in Table~\ref{metric_runtime}.
In practice, we observe that StableSR is capable of generating high-quality results much faster using advanced samplers in fewer sampling steps.
Specifically, DDIM~\citep{song2020denoising} enables StableSR to generate results with faithful details in 20 steps.
Moreover, StableSR can be further applied to SD-turbo~\citep{sauer2023adversarial} w/o further finetuning.
As shown in Fig.~\ref{fig_turbo}, StableSR equipped with SD-turbo can generate high-quality results with only 4 steps, significantly reducing the inference time, i.e., 0.83s as shown in Table~\ref{metric_runtime}, which is 6.3 times faster than LDM with 200 sampling steps, while still remarkably outperforming popular GAN-based methods~\citep{wang2021realesrgan,liang2021swinir} and LDM~\citep{rombach2021highresolution}.
Notably, directly speeding up LDM using existing fast sampling approaches, i.e., DDIM will lead to a severe performance drop as shown in Fig.~\ref{fig_turbo}.

\section{Limitations}
Though benefiting from the diffusion prior, StableSR also shares similar limitations with it.
Specifically, StableSR may struggle in handling small texts, faces and patterns as shown in Fig.~\ref{fig:failcase}.
While these cases are challenging for existing generic super-resolution approaches including StableSR, we believe adopting a more powerful diffusion prior and training on more high-quality data can help. We leave these as future work.

\vspace{-0.3cm}
\section{Conclusion}
Motivated by the rapid development of diffusion models and their wide applications to downstream tasks, this work discusses an important yet underexplored problem of \textit{how diffusion prior can be adopted for super-resolution.}
In this paper, we present StableSR, a new way to exploit diffusion prior for real-world SR while avoiding source-intensive training from scratch.
We devote our efforts to tackling the well-known problems, such as high computational cost and fixed resolution, and propose respective solutions, including the time-aware encoder, controllable feature wrapping module, and progressive aggregation sampling scheme.
Extensive experiments are conducted for evaluation and effective inference strategies are further provided to facilitate practical applications.
We believe that our exploration would lay a good foundation in this direction, and our proposed StableSR could provide useful insights for future works.

\noindent{\textbf{Acknowledgement:}} This study is supported by the National Research Foundation, Singapore under its AI Singapore Programme (AISG Award No: AISG2-PhD-2022-01-033[T]), RIE2020 Industry Alignment Fund Industry Collaboration Projects (IAF-ICP) Funding Initiative, as well as cash and in-kind contribution from the industry partner(s).
We sincerely thank Yi Li for providing valuable advice and building the WebUI implementation\footnote{https://github.com/pkuliyi2015/sd-webui-stablesr} of our work.
We also thank the continuous interest and contributions from the community.

\appendix
\section*{Appendix}

\section{Details of Time-aware Encoder}
As mentioned in the main paper, the architecture of the time-aware encoder is similar to the contracting path of the denoising U-Net in Stable Diffusion~\citep{rombach2021highresolution} with much fewer parameters (${\sim}$105M, including SFT layers) by reducing the number of channels.
The detailed settings are listed in Table \ref{hyper_table}.

\begin{table}[htbp]
    \centering
    \caption{Settings of the time-aware encoder in StableSR.}
	\label{hyper_table}
	\begin{tabular}{cc}
		\toprule
		\textit{Settings} & \textit{Value}\\
        \midrule
        in\_channels & 4 \\
        model\_channels & 256 \\
        out\_channels & 256 \\
        num\_res\_blocks & 2 \\
        dropout & 0 \\
        channel\_mult & [1, 1, 2, 2] \\
        attention\_resolutions & [4, 2, 1] \\
        conv\_resample & True \\
        dims & 2 \\
        use\_fp16 & False \\
        num\_heads & 4 \\
	\bottomrule
	\end{tabular}
\end{table}

\begin{figure}[htbp]
\begin{center}
\centerline{\includegraphics[width=\columnwidth]{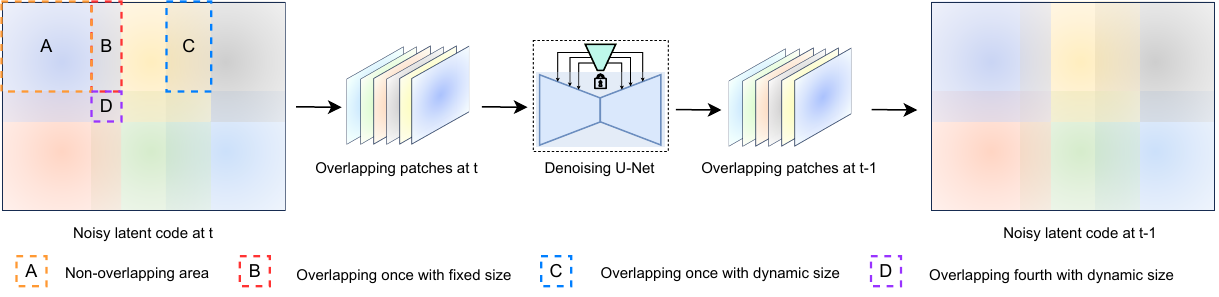}}
\caption{Illustration of our aggregation sampling algorithm. We divide the noisy latent codes into overlapping patches and fuse these patches using a Gaussian kernel at each diffusion iteration. To avoid altering the output resolution, the overlapping size (region C) at the right and bottom boundaries is dynamically adjusted to fit the target resolution.}
\label{fig:aggsampling}
\end{center}
\vspace{-1.5cm}
\end{figure}

\section{Aggregation Sampling}
Here, we provide more details about our aggregation sampling strategy, which is an effective and practical solution that enables arbitrary-size image generation without a perceptible performance drop for diffusion-based restoration.
Our aggregation sampling strategy is mainly inspired by Jim\'{e}nez~\citep{jimenez2023mixture} and we further enable more flexible resolution by dynamically adjusting the overlapping size at the right and bottom boundaries as shown in Fig.~\ref{fig:aggsampling}.

\begin{figure}[htbp]
\begin{center}
\vspace{-0.5cm}
\centerline{\includegraphics[width=1.0\columnwidth]{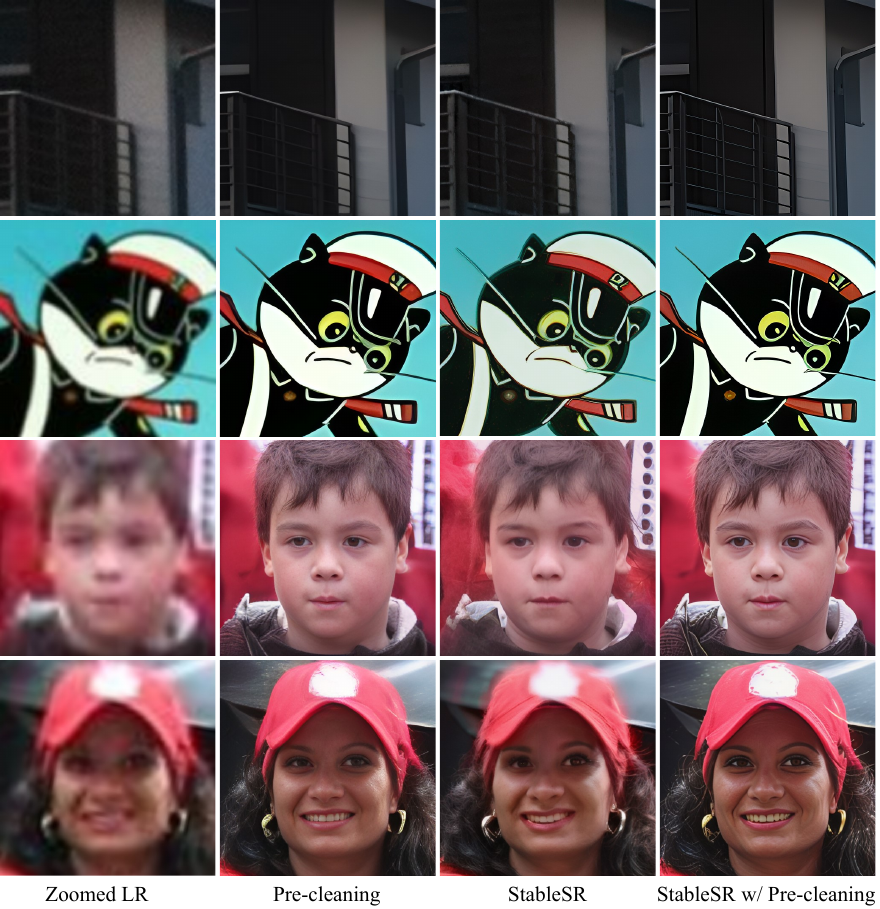}}
\caption{StableSR may generate suboptimal results when the inputs have severe degradations. Adopting a simple pre-cleaning with a pre-trained SR model during sampling can effectively improve the performance of StableSR under such circumstances.
}
\label{fig_preclean}
\end{center}
\vspace{-1.5cm}
\end{figure}

\section{Pre-cleaning for Severe Degradations}
It is observed that StableSR may yield suboptimal results when LR images are severely degraded with pronounced levels of blur or noise, as shown in the first column of Fig. \ref{fig_preclean}. 
Drawing inspiration from RealBasicVSR \citep{chan2022investigating}, we incorporate an auxiliary pre-cleaning phase preceding StableSR to address scenarios under severe degradations.
Specifically, we first adopt an existing SR approach \eg, Real-ESRGAN+ \citep{wang2021realesrgan} for general SR and CodeFormer \citep{zhou2022codeformer} for face SR\footnote{For face SR, we further finetune our StableSR model for 50 epochs on FFHQ \citep{karras2019style} using the same degradations as CodeFormer \citep{zhou2022codeformer}.} to mitigate the aforementioned severe degradations. 
To suppress the amplification of artifacts originating from the pre-cleaning phase, a subsequent $2\times$ bicubic downsampling operation is further adopted after pre-cleaning.
Subsequently, StableSR is used to generate the final outputs.
As shown in Fig.~\ref{fig_preclean}, such a pre-cleaning stage substantially improves the robustness of StableSR.

\section{Additional Visual Results}

\begin{figure*}[!t]
	\begin{center}
		\centerline{\includegraphics[width=2.0\columnwidth]{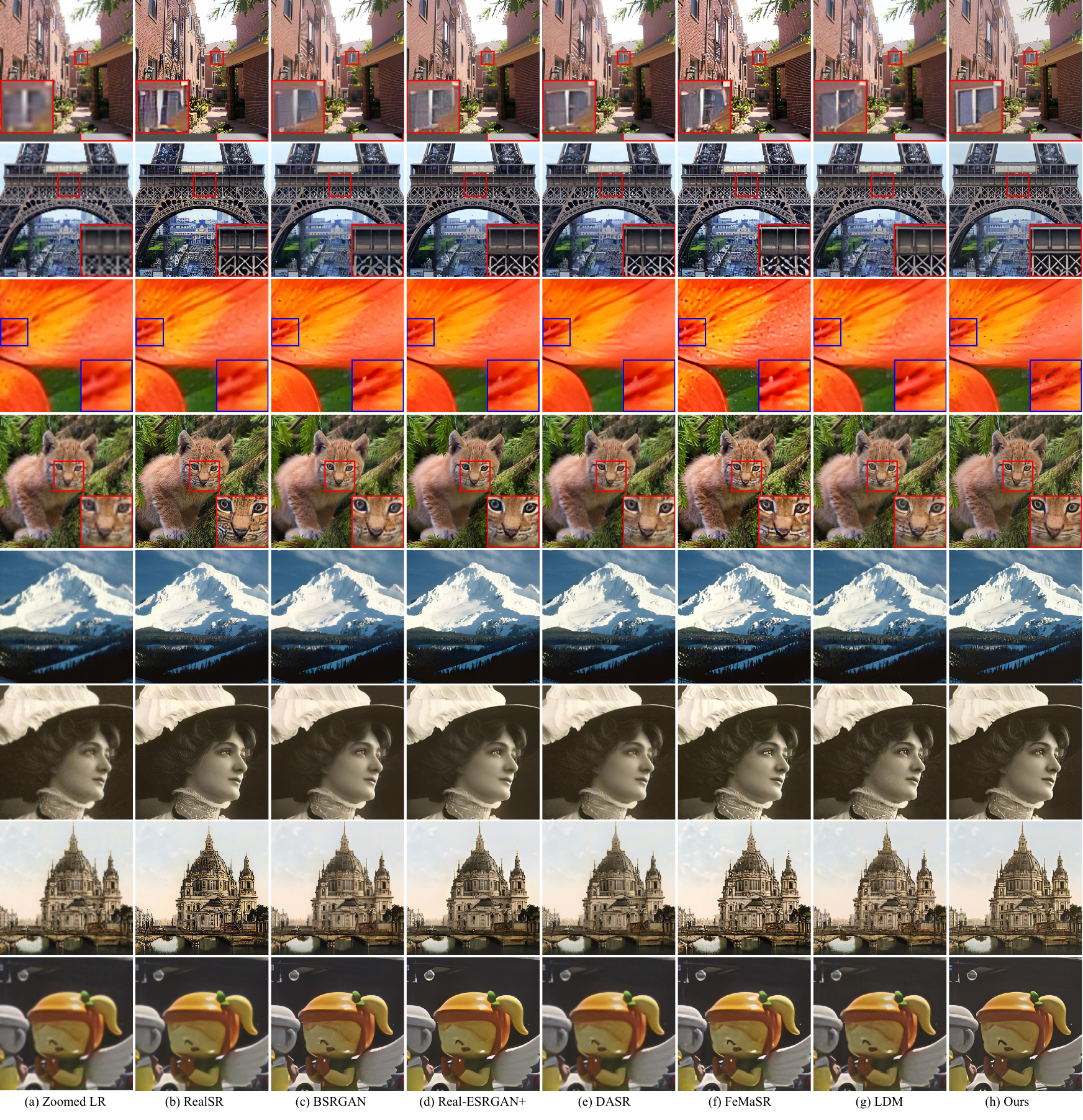}}
		\caption{More qualitative comparisons on real-world images ($128 \rightarrow 512$). While existing methods typically fail to restore realistic textures under complicated degradations, our StableSR outperforms these methods by a large margin. (\textbf{Zoom in for details})}
	\label{fig_fix}
	\end{center}
\end{figure*}

\subsection{Visual Results on Fixed Resolution}
In this section, we provide additional qualitative comparisons on real-world images w/o ground truths under the resolution of $512 \times 512$.
We obtain LR images with $128 \times 128$ resolution.
As shown in Fig. \ref{fig_fix}, StableSR successfully produces outputs with finer details and sharper edges, significantly outperforming state-of-the-art methods.

\subsection{Visual Results on Arbitrary Resolution}
\begin{figure*}[!t]
	\begin{center}
		\centerline{\includegraphics[width=2.0\columnwidth]{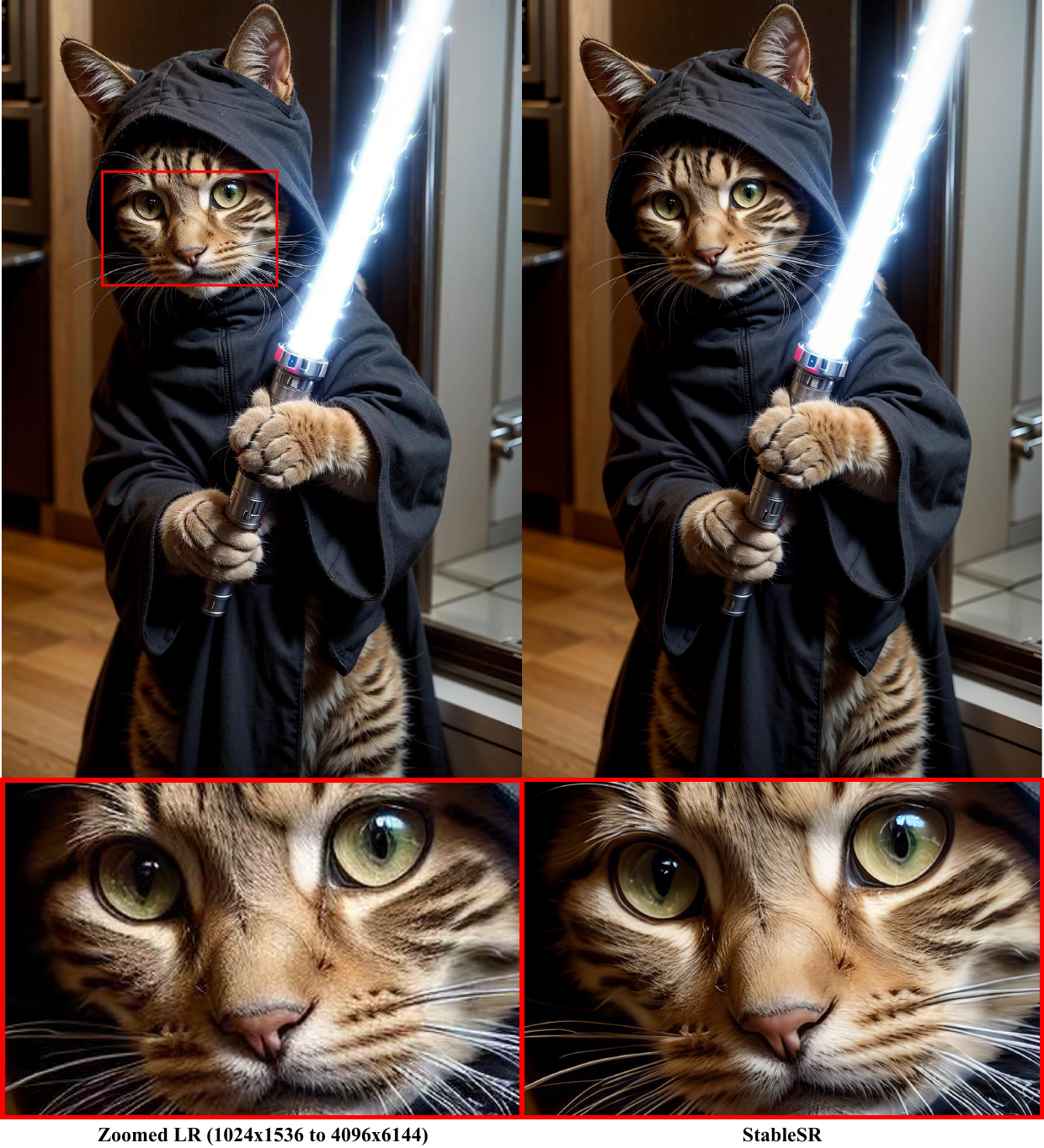}}
		\caption{A 4x StableSR result on AIGC content beyond 4K resolution. (\textbf{Zoom in for details})}
	\label{fig_4k}
	\end{center}
\end{figure*}

\begin{figure*}[!t]
	\begin{center}
		\centerline{\includegraphics[width=2.0\columnwidth]{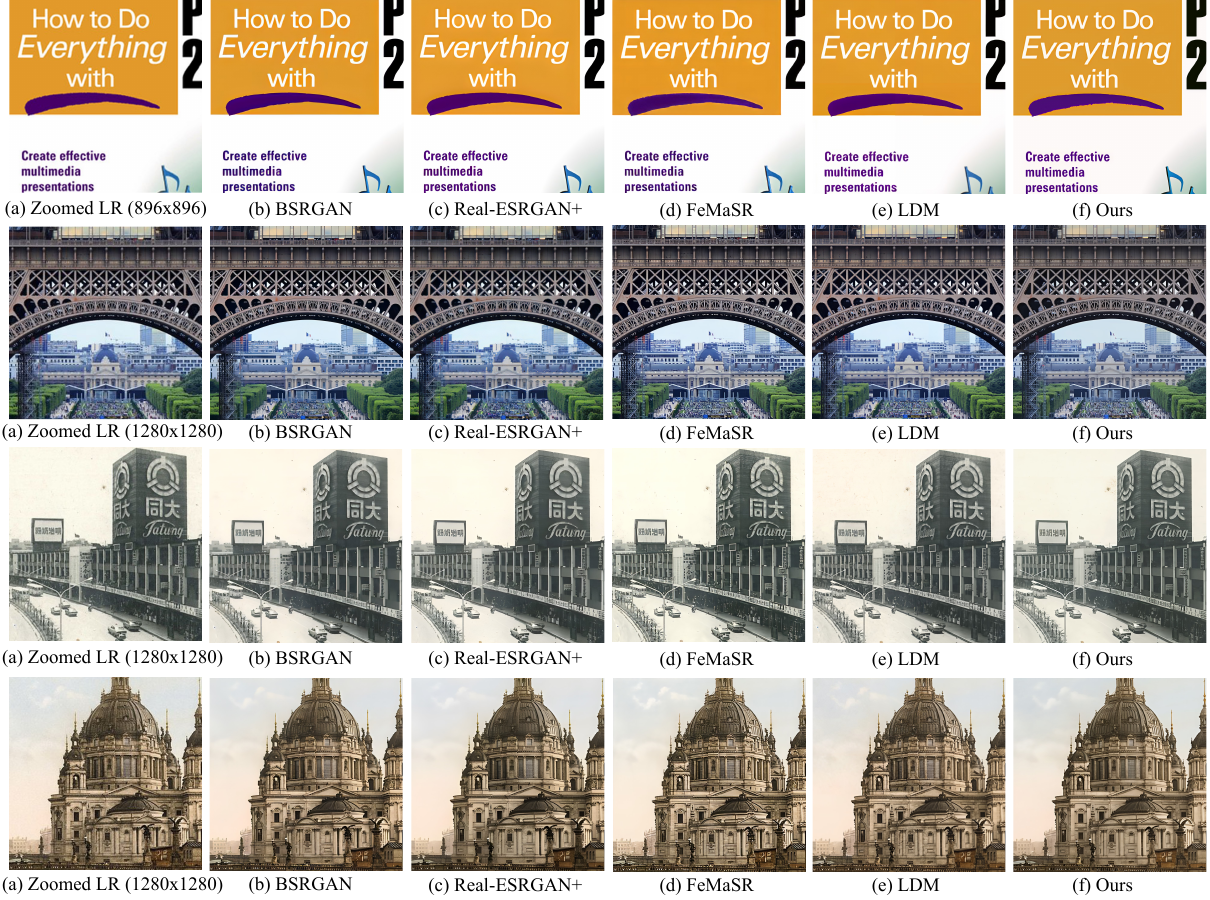}}
		\caption{More qualitative comparisons on original real-world images with diverse resolutions. Our StableSR is capable of generating vivid details without annoying artifacts. (\textbf{Zoom in for details})}
	\label{fig_arb_sup}
	\end{center}
\end{figure*}

In this section, we provide additional qualitative comparisons on the original resolution of real-world images w/o ground truths.
As shown in Fig. \ref{fig_4k}, StableSR is capable of generating high-quality SR images beyond 4x resolution, indicating its practical use in real-world applications.
Moreover, the results in Fig. \ref{fig_arb_sup} indicate that StableSR can generate realistic textures under diverse and complicated real-world scenarios such as buildings and texts, while existing methods either lead to blurry results or introduce unpleasant artifacts.

\bibliographystyle{spbasic}      
\bibliography{ref}   
\end{document}